\documentclass[nohyperref]{article}

\usepackage{microtype}
\usepackage{graphicx}
\usepackage{subfigure}
\usepackage{booktabs} 

\usepackage{hyperref}

\usepackage[accepted]{icml2022}

\usepackage{amsmath}
\usepackage{amssymb}
\usepackage{mathtools}
\usepackage{amsthm}

\usepackage{makecell}
\usepackage{graphicx}
\usepackage{hyperref}
\usepackage{multirow}
\usepackage{wrapfig}
\usepackage{booktabs}
\usepackage{url}
\usepackage{enumitem}
\usepackage{soul}
\usepackage{amsmath}

\usepackage[capitalize,noabbrev]{cleveref}

\theoremstyle{plain}

\theoremstyle{definition}

\theoremstyle{remark}

\usepackage[textsize=tiny]{todonotes}
\usepackage{colortbl}

\icmltitlerunning{3D Infomax}
\definecolor{bad}{HTML}{DB4325}
\definecolor{slight-bad}{HTML}{ffa861}
\definecolor{bit-bad}{HTML}{ffd2ad}
\definecolor{neutral}{HTML}{f4f4f4}
\definecolor{neutral-text}{HTML}{9c9c9c}
\definecolor{bit-good}{HTML}{a5d0ee}
\definecolor{slight-good}{HTML}{63aee2}
\definecolor{textorange}{HTML}{fe8216}
\definecolor{textblue}{HTML}{2178b4}
\definecolor{good}{HTML}{006164}
\definecolor{darkgreen}{rgb}{0.0, 0.5, 0.0}
\definecolor{mplorange}{HTML}{fe8216}
\definecolor{mplblue}{HTML}{2178b4}

\newcommand{\col}[1]{\textcolor{darkgreen}{#1}}
\renewcommand{\col}[1]{#1}
\begin{document}

\twocolumn[
\icmltitle{3D Infomax improves GNNs for Molecular Property Prediction}



\icmlsetsymbol{equal}{*}

\begin{icmlauthorlist}
\icmlauthor{Hannes Stärk}{equal,sch}
\icmlauthor{Dominique Beaini}{comp}
\icmlauthor{Gabriele Corso}{sch}
\icmlauthor{Prudencio Tossou}{comp}
\icmlauthor{Christian Dallago}{yyy}
\icmlauthor{Stephan Günnemann}{yyy}
\icmlauthor{Pietro Liò}{cam}

\end{icmlauthorlist}

\icmlaffiliation{yyy}{Department of Informatics, Technical University of Munich, DE}
\icmlaffiliation{comp}{Valence Discovery, Montreal, CA}
\icmlaffiliation{cam}{Department of Computer Science and Technology, University of Cambridge, UK}
\icmlaffiliation{sch}{EECS, Massachusetts Institute of Technology, Cambridge MA, USA}

\icmlcorrespondingauthor{Hannes Stärk}{hstark@mit.edu}

\icmlkeywords{Machine Learning, ICML}

\vskip 0.3in
]



\printAffiliationsAndNotice{\icmlEqualContribution} 

\begin{abstract}

Molecular property prediction is one of the fastest-growing applications of deep learning with critical real-world impacts.
However, although the 3D molecular graph structure is necessary for models to achieve strong performance on many tasks, it is infeasible to obtain 3D structures at the scale required by many real-world applications.
To tackle this issue, we propose to use existing 3D molecular datasets to pre-train a model to reason about the geometry of molecules given only their 2D molecular graphs.
Our method, called 3D Infomax, maximizes the mutual information between learned 3D summary vectors and the representations of a graph neural network (GNN). 
During fine-tuning on molecules with unknown geometry, the GNN is still able to produce implicit 3D information and uses it for downstream tasks.
We show that 3D Infomax provides significant improvements for a wide range of properties, including a 22\% average MAE reduction on QM9 quantum mechanical properties. Moreover, the learned representations can be effectively transferred between datasets in different molecular spaces.

\end{abstract}

\section{Introduction}
The understanding of molecular and quantum chemistry is a rapidly growing area for deep learning, with models having direct real-world impacts in quantum chemistry \cite{quantum_chemistry}, protein structure prediction \cite{ alphafold2}, materials science \cite{Schmidt2019}, and drug discovery \cite{antibiotic}. In particular, for the task of molecular property prediction, GNNs have had great success \cite{chemprop}. 

GNNs operate on the molecular graph by updating each atom's representation based on the atoms connected to it via covalent bonds. However, these models reason poorly about other important interatomic forces that depend on the atoms' relative positions in space. 
Previous works showed that using the atoms' 3D positions improves the accuracy of molecular property prediction \cite{SchNet, klicpera_dimenet_2020, sphericalmessagepassing2021, klicpera2021gemnet}.

\begin{figure*}[ht]
\center
\includegraphics[width=\textwidth]{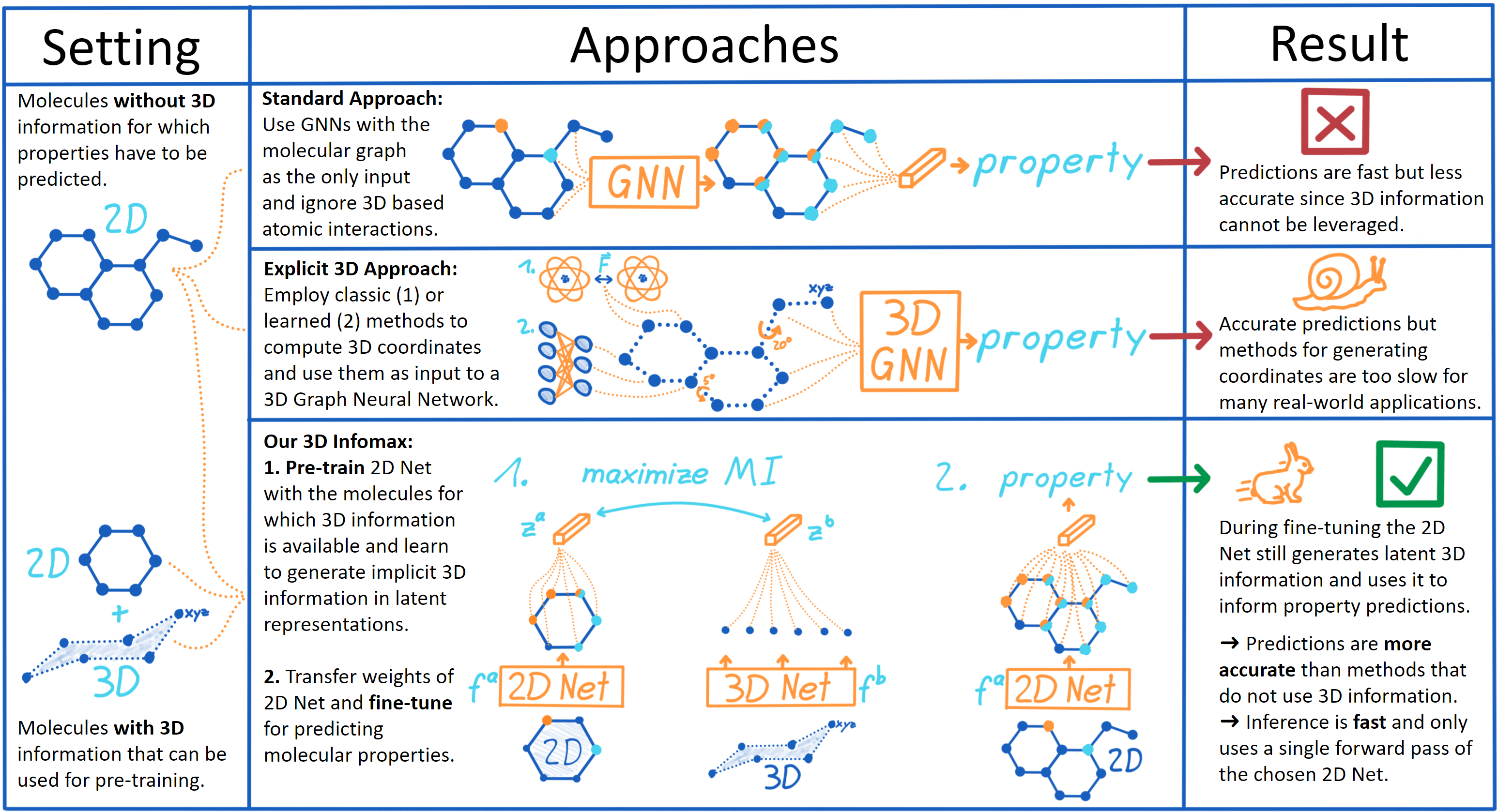}
\caption{The different approaches to molecular property prediction and the motivation for our 3D Infomax pre-training.}
\label{visual-abstract}
\end{figure*}

However, using classical molecular dynamics simulations to explicitly compute a molecule's geometry before predicting its properties is computationally intractable for many real-world applications. Even recent ML methods for conformation generation \cite{LearningGenerativeDyn_MinkaiXu,LearningGradFields,ganea2021geomol} are still too slow for large-scale applications. This issue, as it is summarized in Figure \ref{visual-abstract}, is the motivation for our method.

\textbf{Our Solution: 3D Infomax.}
We propose to pre-train a GNN to encode implicit 3D information in its latent vectors using publicly available molecular structures. In particular, our method, 3D Infomax, pretrains a GNN by maximizing the mutual information \col{(MI)} between its embedding of a 2D molecular graph and a learned representation of the 3D graph. 
This way, the GNN learns to embed latent 3D information using only the information given by the 2D molecular graphs.
After pre-training, the weights can be transferred and fine-tuned on molecular tasks where no 3D information is available. For those molecules, the GNN is still able to produce implicit 3D information that can be used to inform property predictions.

Several other self-supervised learning (SSL) methods that do not use 3D information have been proposed and evaluated to pre-train GNNs and obtain better property predictions after fine-tuning \cite{strategies_for_pretraining_gnn_hu_2020, GraphCL, minghao}. 
These often rely on augmentations (such as removing atoms) that significantly alter the molecules while assuming that their properties \col{do not change. Meanwhile, 3D Infomax pre-training} teaches the model to reason about \col{how atoms interact in space}, which is a principled and generalizable form of information.

We analyze our method's performance by pre-training with multiple 3D datasets before fine-tuning on quantum mechanical properties and evaluating the generalization abilities of the learned representations. 
3D Infomax improves property predictions by large margins and the learned representations are highly generalizable: significant improvements are obtained even when the molecular space of the pre-training dataset is vastly different (e.g., in size) from the kinds of molecules in the downstream tasks.
Moreover, while conventional pre-training methods sometimes suffer from negative transfer \cite{negative_transfer}, i.e., a decrease in performance, this is not observed for 3D Infomax.

\textbf{Our main contributions are:}
\begin{itemize}
    \item A 3D pre-training method that enables GNNs to reason about the geometry of molecules given only their 2D molecular graphs, which improves predictions.
    \item Experiments showing that our learned representations are meaningful for various molecular tasks, without negative transfer.
    \item Empirical evidence that the embeddings generalize across different molecular spaces.
    \item An approach to leverage information from multiple conformers of the same molecule that further improves downstream property predictions and an evaluation to what extent this is possible.
\end{itemize}

\section{Background}

\textbf{2D Molecular Graphs}
A molecule's 2D information can be represented as a graph $G=(\mathcal{V}, \mathcal{E})$ with atoms $\mathcal{V}$ as nodes, and the edges $\mathcal{E}$ given by covalent bonds. The 2D information of edges could contain the bond type while nodes are attributed with features such as the atomic number, but no 3D coordinates. 

\textbf{3D Molecular Conformers.}\label{conformers}
Molecules are dynamic 3D structures that can exist in different spatial conformations. For a single 2D molecular graph, there are multiple low energy atom arrangements that are likely to occur in nature. These are called conformers and they can exhibit different chemical properties. For a model to properly capture 3D information, it is important to consider all the most likely conformations.

When considering a number $c$ of known conformers of a molecule, we represent them as a set of point clouds $\{R^j\}_{j\in\{1\dots c \}}$. Each point cloud $R = \{r_v\}_{v\in \mathcal{V}}$ specifies the locations of all atoms $\mathcal{V}$ in the molecule.

Several tools exist to compute conformers ranging from methods based on classical force fields to slower but more accurate molecular dynamics simulations. Methods such as RDKit's ETKDG algorithm \cite{rdkit} are fast but less accurate. The popular metadynamics method CREST \cite{crest} offers a good tradeoff between speed and accuracy but still requires about 6 hours per drug-like molecule per CPU-core \cite{axelrod2020geom}. This makes it infeasible to explicitly compute precise structures in applications where large numbers of molecules need to be processed. For instance, virtual screening for drug discovery where screening datasets sometimes are comprised of millions or billions of molecules \cite{enamine_real_database}.


\textbf{Symmetries of Molecules.}
A molecule's conformation does not change if all the atom coordinates are jointly translated or rotated around a point, i.e., molecules are symmetric with respect to these \col{two types of transformations which is also known as SE(3) symmetry}. Note that some molecules (called chiral) are not invariant to reflections: their properties depend on their chirality. Deep learning architectures that capture these symmetries are usually more sample efficient, and they generalize to all symmetric inputs the architecture has been designed for \cite{5G}. In our method, the produced representations of the 3D structure respect these symmetries.



\textbf{Graph Neural Networks.}\label{graph-neural-networks}
We make use of GNNs to predict molecular properties given a molecular graph. Many GNNs can be described in the framework of MPNNs \cite{mpnn}, such as the PNA model, \cite{pna} which we employ. 

The aim of MPNNs is to learn a representation of a graph $G=(\mathcal{V},\mathcal{E})$ with nodes $\mathcal{V}$ connected by edges $\mathcal{E}$. They do so by iteratively applying message-passing layers and then combining all node representations in a readout function. A message-passing layer updates the representation of a node given its neighbors and the edges between them using permutation invariant functions such as mean, max, or sum. 
After the message-passing layers, another permutation invariant function can be used as a readout to obtain a final graph level representation from the node level embeddings.

\section{Related Work}
\textbf{Molecular property prediction.}
Since \citet{mpnn} introduced the MPNN framework, GNNs became popular for quantum chemistry \cite{GNNFilm,Tang, mpnnForPysicalChemical}, drug discovery \cite{GRLforDrugDIscorver, antibiotic,drug_target_GCN}, and molecular property prediction in general \cite{property_predGCNBarzilay, molecularPropPredCompositionalNets, PhysNet}. The field is well established with easily accessible molecular datasets driving progress \cite{moleculenet, ogb2020hu} and rigorous evaluations of MPNNs for property prediction \cite{chemprop} showing the effectiveness of the approach.

While these GNNs have had great successes by operating on the 2D graph, many tasks on molecules can be improved by additionally using 3D information. A simple approach is to use bond lengths as edge features \cite{chen2020utilizing}, but methods that capture more molecular geometry improve on this such as SchNet \cite{SchNet}. Similarly, DimeNet \cite{klicpera_dimenet_2020, klicpera_dimenetpp_2020} proposed extracting more 3D information via bond angles, which further improved quantum mechanical property prediction. SMP \cite{sphericalmessagepassing2021} included another angular quantity, and GemNet \cite{klicpera2021gemnet} developed an approach to also capture torsion angles, such that all relative atom positions are uniquely defined. EGNN \cite{egnn} achieved the same by operating on all pairwise atom distances.

\textbf{Self-Supervised Learning.}\label{ssl-related-work} (SSL) attempts to find supervision signals in unlabelled data to learn meaningful representations. Contrastive learning \cite{CPC,NCE,MINE,DIM} is a popular class of methods that learn representations by comparing the embeddings of similar and dissimilar inputs and have achieved impressive results in computer vision \cite{SimCLR,SwAV}.

Learning from unlabeled data also is a critical challenge in molecular chemistry since datasets are relatively small due to experimental or computational costs. Several works have explored contrastive learning variants in the context of molecular graphs for non-quantum molecular properties \cite{strategies_for_pretraining_gnn_hu_2020, MolCLR, GraphCL, GraphCLautomated, minghao, ZhuDualView}. However, the improvements these methods provide in molecular property prediction are still limited and often fail to generalize. 

Previous methods for SSL on molecules only leveraged the 2D information of molecules. Meanwhile, 3D Infomax and the concurrently developed GraphMVP \cite{liu2022pretraining} additionally make use of molecules' 3D structures to obtain more informative representations. GraphMVP proposes a generative and a contrastive 3D pre-training task. The generative task can incorporate the information of multiple molecular conformers and adding it to the contrastive pre-training improves downstream performance.  Our work differs from GraphMVP in multiple ways. 3D Infomax does not require an additional generative pre-training task to leverage multiple 3D conformers. Instead, we directly include this information in a new contrastive loss formulation. Next, we use multiple 3D datasets for pre-training that belong to different chemical spaces, which allow us to demonstrate the ability to effectively transfer representations between them. Moreover, our evaluation includes quantum mechanical tasks, and we find that the possible improvements in this domain are much larger than for non-quantum properties. 

\section{3D Infomax}\label{method}
\begin{figure*}[htpb]
  \centering
  \includegraphics[width=0.8\textwidth]{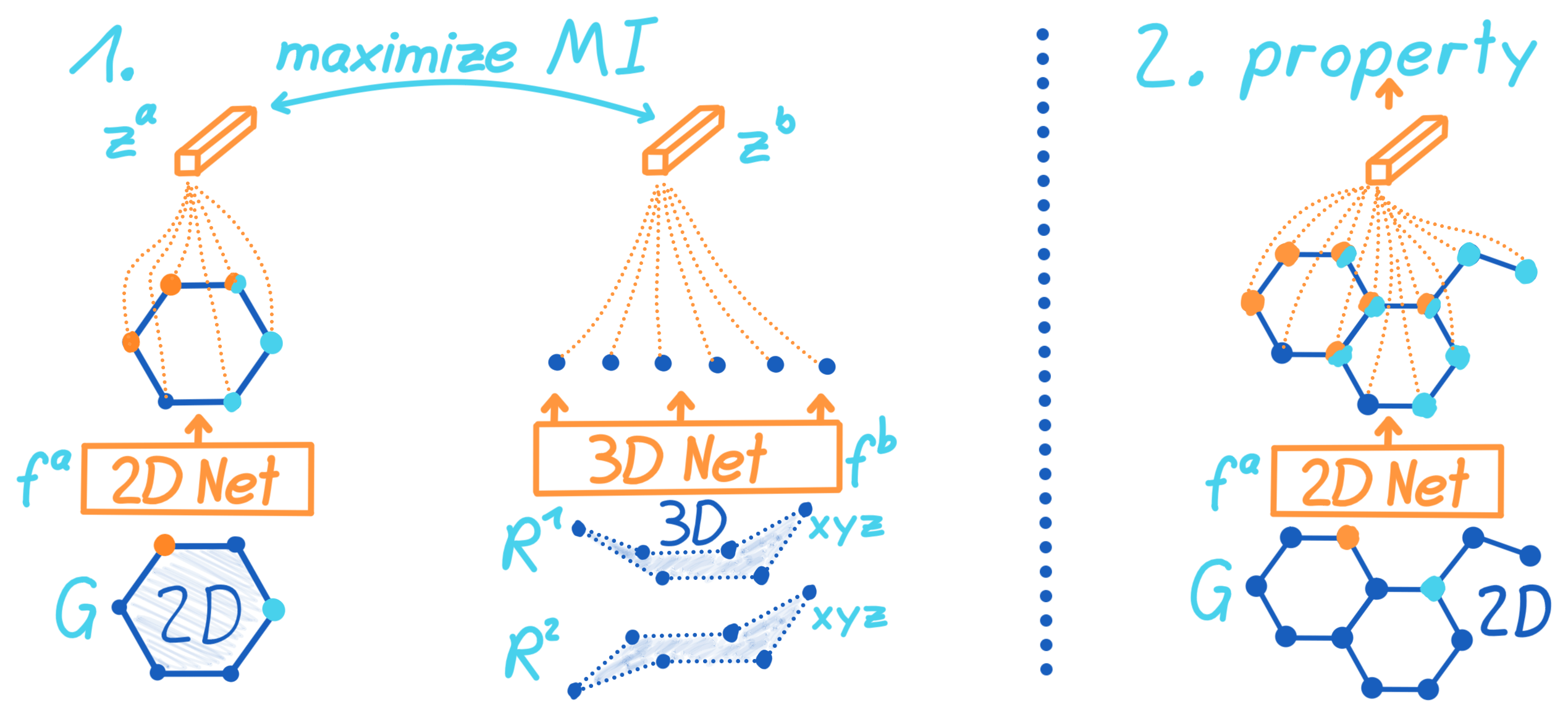}
  \vspace{-0.2cm}
  \caption{We first pre-train a 2D network $f^a$ by maximizing the mutual information (MI) between its representation $z^a$ of a molecular graph $G$ and a 3D representation $z^b$ produced from the molecules' conformers $R^j$. In step 2, the weights of $f^a$ are transferred and fine-tuned to predict properties.}\label{whole-setup-image}
\end{figure*}
To achieve our goal of having a 2D GNN that is able to reason about 3D geometry from only 2D inputs, we pre-train our model using contrastive learning. Figure \ref{whole-setup-image} visualizes our method which maximizes the mutual information between a 2D GNN using the 2D molecular graph and a 3D GNN using the associated 3D conformers. After pre-training, we transfer the weights and fine-tune them on property prediction tasks. During fine-tuning, the GNN's produced 3D information can be used to improve predictions.


3D Infomax uses two different models, as visualized in Figure \ref{whole-setup-image}. Firstly, the model that should be pre-trained which we call \textit{2D network} $f^a$ since its inputs are 2D molecular graphs $G=(\mathcal{V}, \mathcal{E})$ with atoms $\mathcal{V}$ and bonds $\mathcal{E}$ from which it produces a representation $f^a(G)=z^a \in \mathbb{R}^{d_z}$. This can be any GNN that one chooses for the downstream task. 

Secondly, the \textit{3D network} $f^b$ which encodes the atoms' 3D coordinates $R = \{r_v\}_{v\in \mathcal{V}}$ in a 3D representation $f^b(R)=z^b  \in \mathbb{R}^{d_z}$. \col{Our pre-training can also be understood from a contrastive distillation \cite{distillation2019Tian} perspective where the student 2D network learns from the teacher 3D network to produce 3D information.}

\subsection{Contrastive Framework} 
\begin{figure*}[htpb]
  \centering
  \includegraphics[width=0.84\textwidth]{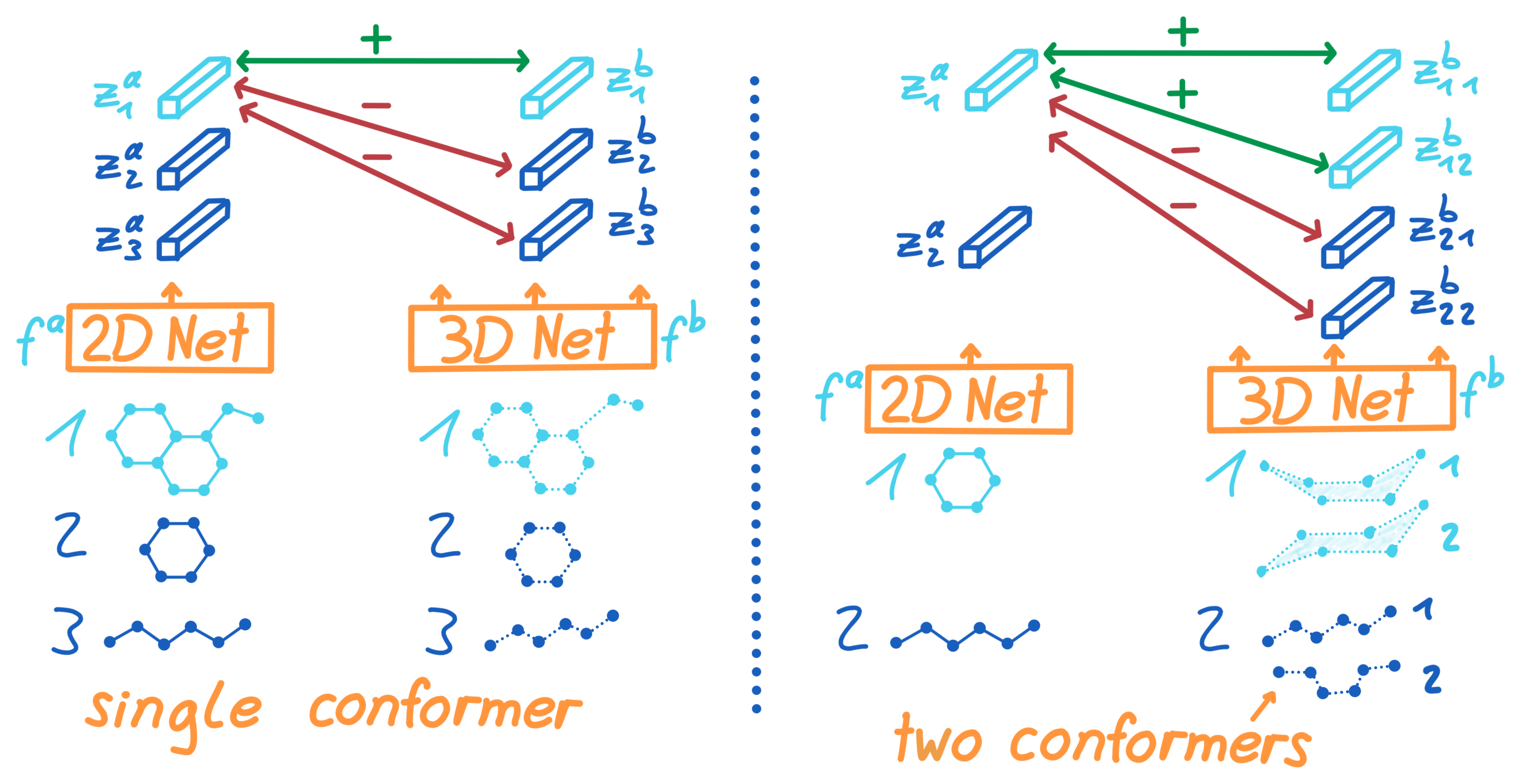}
  \caption{The \textbf{single conformer} example shows a batch of three molecular graphs as input to the 2D network with the corresponding three conformers as input to the 3D model. During 3D pre-training, the contrastive loss $\mathcal{L}$ enforces high similarity between latent representations that come from the same molecule \textcolor{darkgreen}{(green arrows)} while encouraging dissimilarity otherwise \textcolor{red}{(red arrows)}. This is depicted for the first molecule, but the same is calculated for the second and third. The final loss is the average.
  The \textbf{multiple conformer} example on the right shows two conformers per molecule $c=2$, and the loss is adjusted to treat all of them as positive pairs if they come from the same molecule and as negative pairs otherwise. Our loss \col{$\mathcal{L}^{multi3D}$} achieves this.} \label{contrastive-learning-multi3D-image}
\end{figure*}

To teach the 2D network $f^a$ to produce 3D information from the 2D graph inputs, we maximize the mutual information between the latent 2D representations $z^a$ and 3D representations $z^b$.
Intuitively, we wish to maximize the agreement between $z^a$ and $z^b$ if they are derived from the same molecule. For this purpose, we use contrastive learning (visualized in Figure \ref{contrastive-learning-multi3D-image}). We consider a batch of $N$ molecular graphs $\{G_i\}_{i\in\{1\dots N \}}$ with their atom coordinates $\{R_i\}_{i\in\{1\dots N \}}$ from which the networks produce multiple representations $z^a_i$ and $z^b_i$. 

The first objective of contrastive learning is to maximize the representations' similarity if they are a positive pair, meaning that they come from the same molecule (same index $i$).
The second objective is to enforce dissimilarity between negative pairs $z^a_i$ and $z^b_k$ where $i \neq k$, i.e., the 2D and 3D representations in the batch should be dissimilar if they come from different molecules. These objectives are captured in the popular NTXent \col{loss \cite{SimCLR} and we use a similar loss} to jointly optimize our models:
\begin{equation}
    \col{\mathcal{L}} = -\frac{1}{N}\sum_{i=1}^{N}\left[\log\, \frac{e^{sim(z^a_i,\,z^b_i)/\tau}}{\sum_{\substack{k=1 \\ k\neq i}}^{N}e^{sim(z^a_i,\,z^b_k)/\tau}} \right]
\end{equation}
where $sim(z^a, z^b)= z^a \cdot z^b/(\lVert z^a \rVert \lVert z^b \rVert)$ is the cosine similarity and $\tau$ is a temperature parameter which can be seen \col{as weight for the most similar negative pair.} While different combinations of contrastive losses and SSL are possible to learn a joint embedding space between 2D and 3D representations, we found \col{the above loss} to perform best. Other methods (Barlow Twins \cite{BarlowTwins}, BYOL \cite{BYOL}, VICReg \cite{vicreg}) are explored in Appendix \ref{latent-space-ssl}.

\subsection{Using Multiple Conformers}\label{multiple-conformers}
For most molecules, there are multiple low-energy stable conformers. Instead of only using the most probable conformer (with the lowest energy), we found that leveraging structural information from multiple conformers provides significant benefits. To achieve this, we now consider the $c$ highest probability conformers   $\{R^j_i\}_{j\in\{1\dots c \}}$ of the i-th molecule. If there are fewer than $c$ conformers for a molecule, the lowest energy conformer is repeated. Our choice for the following approach is justified by its good trade-off between simplicity and performance in the comparisons with other possible methods in Appendix \ref{multiple-conformers-appendix}.

For every molecule the 3D network now takes all conformers as input and produces their latent 3D representations $\{z^b_{i,j}\}_{j\in\{1\dots c \}}$. The objective is to maximize the similarity between $z^a_i$ and all conformer representations $z^b_{i,j}$ that stem from the same molecule (see Figure \ref{contrastive-learning-multi3D-image}). As such, we modify \col{our loss} to sum over the similarities of all conformers to obtain \col{the} final loss:
\begin{equation}\label{multi3D-loss}
    \col{\mathcal{L}^{multi3D}} = -\frac{1}{N}\sum_{i=1}^{N}\left[\log\, \frac{\sum_{j=1}^c e^{sim(z^a_i,\,z^b_{i,j})/\tau}}{\sum_{\substack{k=1 \\ k\neq i}}^{N}\sum_{j=1}^c e^{sim(z^a_i,\,z^b_{k,j})/\tau}} \right].
\end{equation}

\subsection{3D Network}
The 3D network takes as input the coordinates of the atoms as a 3D point cloud and has to produce an SE(3) invariant representation vector $z^b$ that encodes as much information as possible about the 3D structure. Note that it does not have access to the 2D information such as atom or bond features; otherwise, the empirical estimate of the mutual information could be increased by both networks encoding this information instead of the desired 3D structure. 

Our concrete architecture encodes the 3D information given by the pairwise Euclidean distances of all atoms. This representation uniquely defines all relative atom positions and is invariant to translation and rotation, as desired. However, like the representations used by many other 3D graph processing architectures \cite{klicpera_dimenet_2020, egnn}, it is also invariant to reflection, which prevents it from distinguishing chiral molecules. Using all pairwise distances also means that the model's complexity is quadratic in the number of atoms, which is feasible for drug-like molecules.

The pairwise distances $d_{uv}$ between atoms $u$ and $v$ are first mapped to a higher dimensional space using sine and cosine functions with high frequencies. Recent work provides theoretical \cite{spectral_bias_of_nn} and empirical \cite{tancik2020fourfeat} evidence that this enables deep networks to better fit data with high-frequency variations. Such high-frequency variation is present in the 3D data we wish to encode since differences in bond lengths are often small. Further, the bond distances are likely the most important since close-by atoms typically have the most relevant interactions.
As such, we use the following mapping $\gamma: \mathbb{R} \mapsto \mathbb{R}^{2F+1}$ with the number of frequencies $F$ set to $4$:
\begin{equation}
\begin{split}
    \gamma(d_{uv}) = \Big(
    d_{uv},\, \sin\big(\frac{d_{uv}}{2^0}\big), \,\cos\big(d_{uv}/2^0\big),\, \dots, \,
    \\
    \sin\big(\frac{d_{uv}}{2^{F-1}}\big),\, \cos\big(\frac{d_{uv}}{}\big)
    \Big).
\end{split}
\end{equation}

The further components can be seen as an MPNN \cite{mpnn} operating on the fully connected graph of a molecule with the encoded distances as edge features and a constant learned vector as node features. 
The message passing layers iteratively encode the 3D information into the node features, which are pooled to produce the 3D representation $z^b$. The differences to standard MPNNs are detailed in Appendix \ref{Net3d-details}. \col{Instead of the presented architecture, a 3D GNN such as SMP \cite{sphericalmessagepassing2021} operating on learned node embeddings could also be used. We validate the improvements provided by our architecture} and $\gamma$ encoding with a set of ablation experiments reported in Appendix \ref{3d-network-appendix}.

\section{Experiments}
\subsection{Pre-training Baselines}\label{pre-training-baselines}
\textbf{Distance Predictor.}
A simpler method to use available 3D structures to pre-train a GNN instead of 3D Infomax is to learn to directly predict all atom distances in the lowest energy conformer. To predict the distance between node $v$ and $u$, we concatenate their representations $h_u$, $h_v \in \mathbb{R}^{d_h}$ that were produced by the GNN and feed them to a multi-layer perceptron (MLP) that produces a single scalar $U: \mathbb{R}^{2d_h} \mapsto \mathbb{R}$. The distance prediction $dist_{uv}$ is then given by 
\begin{equation}\label{distance-equation}
    dist_{uv} = \mathrm{softplus}(U(h_v\mathrel{\Vert}h_u) + U(h_u\mathrel{\Vert}h_v))
\end{equation}
where $\mathrel{\Vert}$ denotes concatenation and $\mathrm{softplus}(x)=\log(1+e^{x})$. The node representations are concatenated in both orders and fed to the MLP to ensure that the function is symmetric. The pre-training loss $f$ is the mean squared error between the predicted and true distances.


\textbf{Conformer Generation.} GeoMol \cite{ganea2021geomol} is the state-of-the-art deep learning method for generating molecular conformations. It is a generative model that produces a distribution of likely 3D structures for a single molecule, thus capturing the information of multiple conformers. Their architecture employs a GNN whose node representations are used to obtain the final distribution. To use their model as a baseline, we use their training process as a pre-training task and then extract the GNN and fine-tune it on the different downstream tasks.

\textbf{GraphCL.}
We compare against the conventional augmentation-based pre-training method GraphCL \cite{GraphCL} \col{with the settings of JOAO \cite{GraphCLautomated} since it outperformed other SSL approaches for multiple molecular tasks.} It uses a common self-supervised objective in which the model has to learn to produce representations that are invariant to augmentations. \col{We use randomly dropping nodes with a ratio of $0.2$ on both branches of the SSL setup since JOAO found this combination of augmentations to work particularly well for molecules.}



\subsection{Data and Evaluation Protocol} 

To evaluate our method, we pre-train using subsets of 3D molecular structure datasets. We then fine-tune the models to predict properties of 2D datasets or of 3D datasets while ignoring their 3D information. It is clear that using a 3D GNN and the explicit 3D information of these datasets that was calculated with expensive quantum simulations would yield the best performance. However, we only use the 2D molecular graphs of the fine-tuning data to simulate the scenario when 3D structures are not available. This allows us to compare the improvements through the implicit 3D information of 3D Infomax pre-trained GNN with the maximum possible improvements represented by using the explicit ground truth 3D conformers with a 3D GNN.

The concrete 3D datasets of which we use subsets for pre-training are: \textbf{QM9}  \cite{Ramakrishnan2014} which contains $134$k small molecules (18 atoms on average) with a single conformer, \textbf{GEOM-Drugs} \cite{axelrod2020geom} with $304$k molecules and \textbf{QMugs} \cite{isert2021qmugs} with $665$k. GEOM-Drugs and QMugs, both consist of larger drug-like molecules (44.4 and 30.6 atoms on average) with multiple conformers. 

For fine-tuning, we predict ten quantum properties of subsets of QM9 and GEOM-Drugs, for which we remove the 3D information. These subsets never have molecules that are contained in the pre-training data, even if the dataset is used to draw pre-training and fine-tuning data from. Additionally, we report results on non-quantum properties of ten OGB \cite{ogb2020hu} datasets in Appendix \ref{non-quantum-experiment}. The atom and bond features we use for each 2D molecular graph are the same as used in OGB, and further details about all used data are in Appendix \ref{data-details}.

We choose PNA \cite{pna} as the GNN to pre-train due to its simplicity and state-of-the-art performance for molecular tasks. The reported confidence intervals are one standard deviation calculated from six random weight initializations, unless stated otherwise. \col{All baselines we compare with use the same GNN as our 3D Infomax method and} all experimental settings are detailed in Appendix \ref{experimental-details-appendix}. Code to 3D pre-train a GNN, to generate molecular fingerprint embeddings, or to reproduce results is available at \url{https://github.com/HannesStark/3DInfomax}.

\subsection{Quantum Mechanical Properties}\label{quantum-properties}

\begin{table*}[thbp]
\caption{\col{MAE for QM9's properties.} \textbf{3D Infomax} is tested with three different pre-training datasets and \textbf{GraphCL} uses a two times larger subset of GEOM-Drugs. \col{\textbf{True 3D SMP}} is a 3D GNN using \col{ground truth} 3D coordinates (hidden from other methods). Details on confidence intervals are in Appendix \ref{experimental-details-appendix}. Colors indicate \textcolor{textblue}{improvement} (lower MAE) or \textcolor{textorange}{worse} performance compared to the randomly initialized (\textbf{Rand Init}) model.
}
\label{qm-table}
\begin{center}
\begin{small}
\begin{sc}
\setlength{\tabcolsep}{4.4pt}
\begin{tabular}{lc|cccc|ccc|c||c}
\toprule
&\multicolumn{1}{c}{} &\multicolumn{4}{c}{Pre-training baselines} &  \multicolumn{3}{c}{ Our 3D Infomax}& \multicolumn{1}{c}{\col{RDKit}} & \textit{\col{True} 3D}\\

         Target  & \multicolumn{1}{c}{Rand Init} & GraphCL& ProPred& DisPred & \multicolumn{1}{c}{ConfGen} & \multicolumn{1}{c}{QM9} & Drugs & \multicolumn{1}{c}{QMugs} & \multicolumn{1}{c}{\col{SMP}} & \multicolumn{1}{c}{\textit{SMP}}   \\    
\midrule
$\mu$       &0.4133\scriptsize$\pm$0.003     &\cellcolor{bit-good}0.3937   & \cellcolor{bit-good}0.3975   & \cellcolor{slight-bad}0.4626  & \cellcolor{bit-bad}0.3940  &\cellcolor{slight-good} \textbf{0.3507}    &\cellcolor{slight-good} \textbf{0.3512}    &\cellcolor{slight-good} 0.3668 &\cellcolor{slight-bad} 0.4344   & 0.0726   \\
$\alpha$    &0.3972\scriptsize$\pm$0.014    &\cellcolor{slight-good}0.3295 &\cellcolor{bit-good}0.3732    & \cellcolor{bit-bad}0.3570     & \cellcolor{bit-bad}0.4219  & \cellcolor{slight-good} 0.3268    &\cellcolor{slight-good} 0.2959    &\cellcolor{slight-good} \textbf{0.2807}   &\cellcolor{slight-good} 0.3020 & 0.1542   \\
homo      &82.10\scriptsize$\pm$0.33      &\cellcolor{bit-good}79.57       & \cellcolor{slight-bad}93.11  & \cellcolor{bit-good}80.58     & \cellcolor{bit-good}79.75  &\cellcolor{slight-good}\textbf{68.96}     &\cellcolor{slight-good} 70.78     &\cellcolor{slight-good} 70.77  &\cellcolor{bit-bad} 82.51    & 56.19  \\
lumo      &85.72\scriptsize$\pm$1.62     & \cellcolor{bit-good}80.81       & \cellcolor{slight-bad}99.84  & \cellcolor{bit-good}84.93     & \cellcolor{bit-good}79.16  &\cellcolor{slight-good} \textbf{69.51}     &\cellcolor{slight-good} 71.38     &\cellcolor{slight-good} 78.10   &\cellcolor{bit-good} 80.36  & 43.58 \\
gap       &123.08\scriptsize$\pm$3.98     &\cellcolor{bit-good}120.08      &\cellcolor{slight-bad}131.99  & \cellcolor{bit-good}116.21    & \cellcolor{bit-good}110.72   &\cellcolor{slight-good} \textbf{101.71}    &\cellcolor{slight-good} 102.59    &\cellcolor{slight-good} 103.85 &\cellcolor{bit-good} 114.24  & 85.10 \\
r2        &22.14\scriptsize$\pm$0.21      &\cellcolor{bit-good}21.84       &\cellcolor{slight-bad}29.21   & \cellcolor{slight-bad}29.23   & \cellcolor{bit-good}20.86      &\cellcolor{slight-good} \textbf{17.39}     &\cellcolor{slight-good} 18.96     &\cellcolor{slight-good} 18.00  &\cellcolor{bit-bad} 22.63   & 1.51   \\
ZPVE      &15.08\scriptsize$\pm$2.83     &\cellcolor{bit-good}12.39        &\cellcolor{bit-good}11.17     & \cellcolor{slight-bad}25.91   & \cellcolor{slight-bad}21.10    &\cellcolor{slight-good} 7.966     &\cellcolor{slight-good} 9.677     & \cellcolor{bit-good}12.06 & \cellcolor{slight-good} \textbf{5.18}   & 2.69 \\
$c_v$       &0.1670\scriptsize$\pm$0.004    & \cellcolor{bit-good}0.1422   &\cellcolor{slight-bad}0.1795  & \cellcolor{bit-good}0.1587    & \cellcolor{bit-good}0.1555    &\cellcolor{slight-good} 0.1306    &\cellcolor{bit-good} 0.1409    &\cellcolor{slight-good} \textbf{0.1208} & \cellcolor{bit-good} 0.1419   & 0.0498  \\ \bottomrule
\end{tabular}
\setlength{\tabcolsep}{6pt}
\end{sc}
\end{small}
\end{center}
\vskip -0.1in
\end{table*}

\textbf{Pre-training setup.} We use 3D Infomax to pre-train three different instances of PNA (1) on $50$k molecules from QM9 using a single conformer, (2) on $140$k of GEOM-Drugs with 5 conformers and, (3) on $620$k of QMugs using 3 conformers. For comparison, we \col{use two different conventional 2D pre-training methods. These are GraphCL \cite{GraphCL} as described in Section \ref{pre-training-baselines} and pre-training by predicting the \col{Gibbs free enery} of GEOM-Drugs' pre-training subset (labeled \textit{ProPred}).} All pre-training methods use a batch size of $500$.

\textbf{Fine-tuning setup.} After pre-training, the models are fine-tuned on $50$k molecules from QM9 \col{(in Table \ref{qm-table})} or $140$k from GEOM-Drugs \col{(in Table \ref{drugs-table})} that have no overlap with the molecules from the pre-training data. On the same molecules, we also train PNA with random weight initialization (labeled \textit{Rand Init}) to compare how much the downstream performance is improved by the different pre-training methods. 

\textbf{Explicit 3D baseline and ground truth comparison setup.} \col{We additionally train and test the 3D GNN SMP \cite{sphericalmessagepassing2021} with 3D coordinates generated by RDKit's ETKDG algorithm, \cite{rdkit} which can be done in a fast manner (labeled \textit{RDKit SMP}). Using conformers generated by the state-of-the-art learned method, GeoMol \cite{ganea2021geomol} always performed worse (Appendix \ref{geomol-conformers}). Lastly, we evaluate SMP using the accurate ground truth 3D conformers of QM9 which were computed with time-consuming simulations that would be infeasible for many real-world applications. These structures are not available to the other methods.}

\textbf{3D Infomax for QM9 vs. conventional pre-training.} Table \ref{qm-table} shows that 3D Infomax pre-training leads to large improvements over the randomly initialized baseline and over GraphCL with all three pre-training datasets. After 3D pre-training on one half of QM9, the \col{average decrease in MAE is 22\%}. Comparing 3D Infomax on GEOM-Drugs with GraphCL shows that even though the latter is pre-trained on two times as many molecules from the same dataset, 3D pre-training is always better by a large margin.

\textbf{Generalization from pre-training to fine-tuning.} Pre-training with the disjoint half of QM9 performs best since it shares the molecular space of the test set. Nevertheless, the learned representations also generalize well: pre-training on GEOM-Drugs and QMugs leads to improvements of \col{19\% and 18\%} respectively, even though QM9 contains much smaller molecules with on average 18 atoms compared to the 44.4 atoms for the drug-like molecules of GEOM-Drugs. 

\textbf{Comparison with Ground Truth Conformers.} 3D Infomax' large improvements have to be compared to methods that also do not use explicit ground truth 3D information and only operate on the 2D molecular graph for the molecules of which properties should be predicted. However, for the quantum property experiment datasets we employ, high-quality ground truth 3D information was calculated with expensive quantum simulations and we are thus able to compare with 3D GNNs that use this additional input for property prediction. 

The results of SMP in Table \ref{quantum-properties} show that for some properties, the MAE of 3D Infomax with its implicit 3D information is still higher than what is possible when using explicit ground truth conformers (which are time-consuming and expensive to obtain). 
One likely contributing factor for this is that QM9's properties are conformer-specific. There might be a maximum accuracy that can be achieved if only the molecule is known and not for which conformer the property should be predicted. Nevertheless, this performance gap suggests that there is still room for improvement.

\textbf{3D pre-training Baselines.} 3D pre-training by directly predicting 3D quantities is simpler than 3D Infomax and would be preferable in case of similar gains. Therefore, we compare with the baselines in Section \ref{pre-training-baselines} using the same $140$k molecules of GEOM-Drugs for all 3D pre-training methods. \textit{DisPred} refers to predicting all atom distances of the highest probability conformer, and \textit{ConfGen} means pre-training by predicting up to 10 conformers.
Table \ref{quantum-properties} shows that 3D Infomax pre-training is always superior to the 3D pre-training baselines and is the only method to not suffer from negative transfer \cite{negative_transfer}.

\begin{table}[thb]
\caption{The MAE for predicting GEOM-Drugs' properties. \textbf{3D Infomax} compared with \textbf{GraphCL} and no pre-training.}
\label{drugs-table}
\begin{center}
\begin{small}
\begin{sc}
\begin{tabular}{lll}
\toprule
Method & Gibbs & $\langle E \rangle$  \\    
\midrule
Rand Init & .2035 &  .1026  \\
GraphCL & .1941 & .0995   \\
3D Infomax QM9 & .1852  & .0968   \\
3D Infomax Drugs & \textbf{.1811} & \textbf{.0952} \\
3D Infomax QMugs & .1835 & .0965  \\ \bottomrule

\end{tabular}
\end{sc}
\end{small}
\end{center}
\vskip -0.1in
\end{table}

\textbf{3D Infomax for GEOM-Drugs.} Table \ref{drugs-table} further confirms that 3D Infomax substantially improves quantum property predictions and generalizes out-of-distribution. Our method outperforms GraphCL, even though GraphCL also sees the fine-tuning molecules during pre-training. Moreover, we observe strong generalization when pre-training with QM9 and fine-tuning on GEOM-Drugs. In this case, the pre-training data only contains the elements C, H, N, O, and F while the target data contains eleven additional elements that are unseen during pre-training.

\textbf{Results interpretation.} Such consistent and out-of-distribution improvements can be explained by the type of information captured with 3D Infomax. Learning to reason about molecular geometry and its impact does not depend on the data's molecular space. Therefore it is not necessary to have a high similarity between the molecules during pre-training and fine-tuning. 

Another advantage of 3D Infomax is its comparably fast convergence. Pre-training on $620$k molecules of QMugs with 3 conformers takes 12 hours, compared to 71 hours for GraphCL on $280$k molecules of GEOM-Drugs.

\subsection{Number of Conformers and pre-training Molecules}\label{number-conformers-experiment}

\begin{figure}[tbh]
  \centering
\includegraphics[width=0.9\columnwidth]{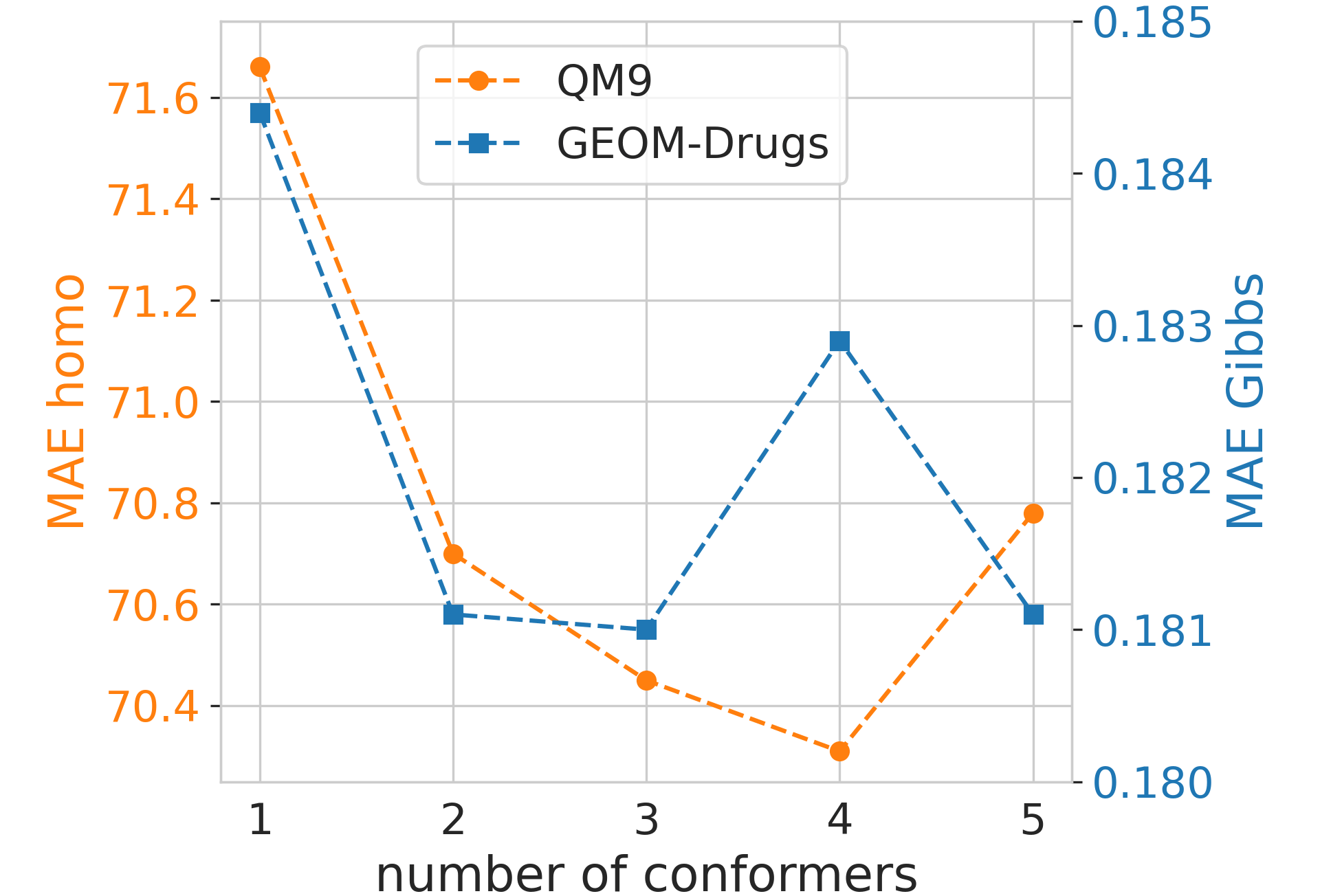}\label{vary-num-conformers-multi3d}
\vspace{-0.1cm}
  \caption{The MAE for the \textcolor{mplorange}{QM9's homo} and \textcolor{mplblue}{GEOM-Drugs' Gibbs} property when varying the number of GEOM-Drugs' conformers used during pre-training.} \label{vary-num-conformers-image}
\end{figure}

\begin{figure}[tbh]
  \centering
\includegraphics[width=0.9\columnwidth]{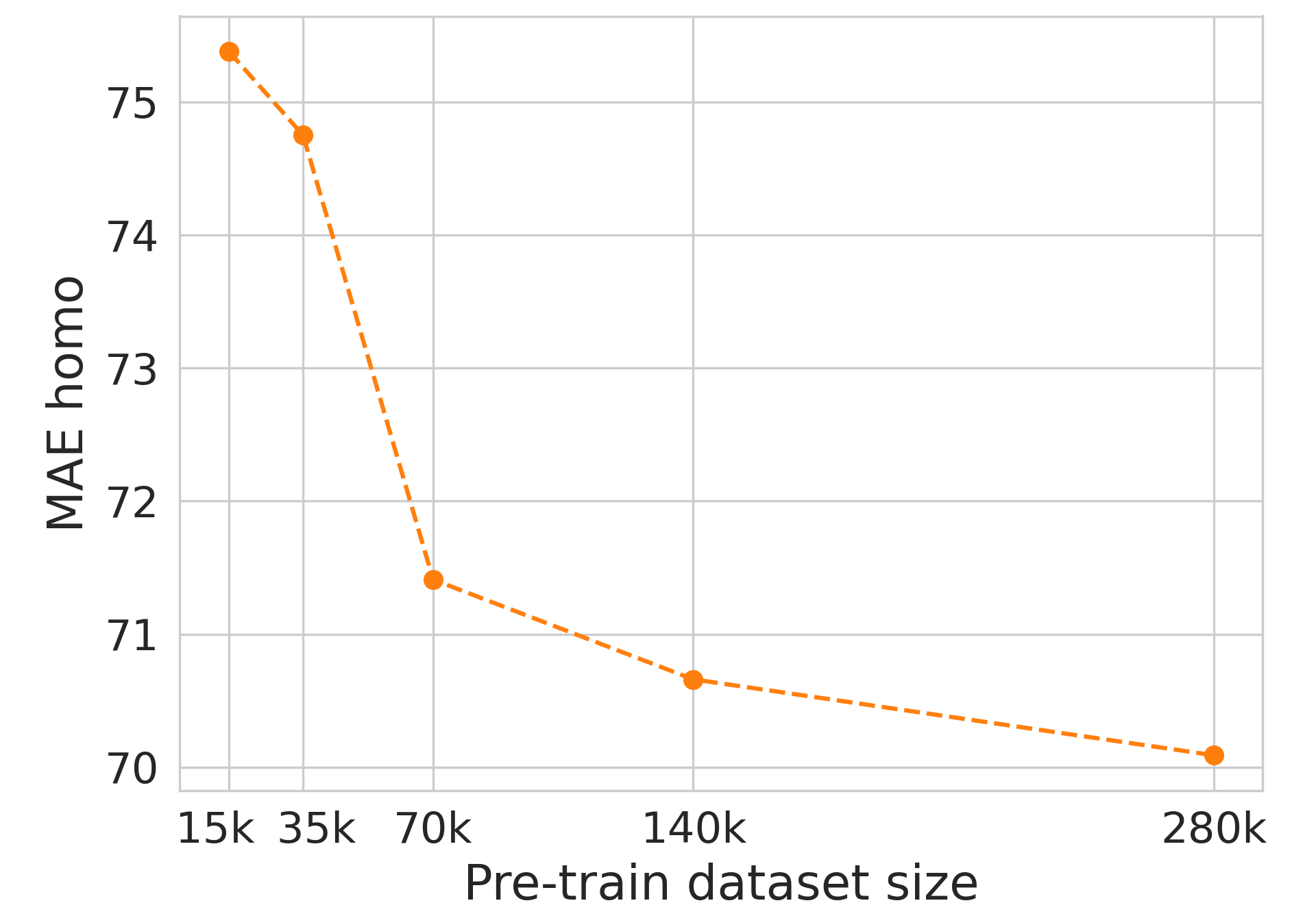}\label{dataset-size-image}
\vspace{-0.1cm}
  \caption{The MAE when using different numbers of molecules of GEOM-Drugs during pre-training.} \label{vary-train-size}
\end{figure}

Figure \ref{vary-num-conformers-image} highlights the benefit of using more than a single conformer. However, the marginal gain reduces as higher energy conformers are added and beyond a certain point (around three conformers), the reduced focus on the most likely conformers worsens the downstream performance. This is in line with the observation that, on average, three conformers are enough to cover 70\% of the cumulative Boltzmann weight for GEOM-Drugs. 
Additionally, experiments in Appendix \ref{multiple-conformers-appendix} show that using multiple conformers is essential when pre-training with QMugs: the MAE for QM9's homo property is $82.57$ with a single conformer while it improves to $70.77$ when using three.

In Figure \ref{vary-train-size} we can observe the performance improving as the size of the pre-training dataset increases. However, the returns are diminishing, and we cannot claim that even larger pre-training datasets are likely to drastically improve performance.

\section{Conclusion}\label{conclusion}
We presented a pre-training strategy, 3D Infomax, that teaches a GNN to produce latent 3D and quantum information from 2D molecular graphs. This can later be used during fine-tuning to improve molecular property predictions while retaining the inference speed of a standard GNN operating on 2D molecular graphs. We found consistently large improvements ({\raise.17ex\hbox{$\scriptstyle\sim$}}22\%) for quantum properties, overshadowing the gains possible with conventional SSL methods. The embedded 3D knowledge can be transferred across highly different types of molecules (e.g., from molecules with an average of 18  atoms to drug-like molecules with 44.4 atoms) since the representations capture a principled form of information that is known to be useful for several molecular tasks. Lastly, we observed that using multiple molecular conformers during pre-training provides valuable additional information to further improve downstream property predictions. 

We provide open-source access to our method at \url{https://github.com/HannesStark/3DInfomax} with a simple setup to 3D pre-train a GNN and the additional possibility to generate molecular fingerprint embeddings that carry latent 3D information from a list of SMILES.

\section{Acknowledgments}
The authors express their gratitude to Simon Axelrod (GEOM Dataset and molecular physics advice), Limei Wang and Meng Liu (Spherical Message Passing), Adrien Bardes and Samuel Lavoie (SSL, VICReg, and BYOL), Octavian Ganea and Lagnajit Pattanaik (molecular
geometry, GeoMol), Christopher Scarveils (part of the project in its beginnings), Bertrand Charpentier and Aleksandar Bojchevski (general help) as well as Zekarias Kefato, Grégoire Mialon, Kenneth Atz, Johannes Klicpera, Minghao Xu, Mozhi Zhang, Shantanu Thakoor, Minkai Xu, Shitong Luo, Yuning You, and Prannay Khosla for insightful discussions.

CD acknowledges support by the Bundesministerium für Bildung und Forschung (BMBF), project number 01IS17049.

\bibliography{example_paper}

\begin{thebibliography}{61}
\providecommand{\natexlab}[1]{#1}
\providecommand{\url}[1]{\texttt{#1}}
\expandafter\ifx\csname urlstyle\endcsname\relax
  \providecommand{\doi}[1]{doi: #1}\else
  \providecommand{\doi}{doi: \begingroup \urlstyle{rm}\Url}\fi

\bibitem[Axelrod \& Gomez-Bombarelli(2020)Axelrod and
  Gomez-Bombarelli]{axelrod2020geom}
Axelrod, S. and Gomez-Bombarelli, R.
\newblock Geom: Energy-annotated molecular conformations for property
  prediction and molecular generation.
\newblock \emph{arXiv preprint arXiv:2006.05531}, 2020.

\bibitem[Axelrod \& G{\'{o}}mez{-}Bombarelli(2020)Axelrod and
  G{\'{o}}mez{-}Bombarelli]{conformerEnsembles}
Axelrod, S. and G{\'{o}}mez{-}Bombarelli, R.
\newblock Molecular machine learning with conformer ensembles.
\newblock \emph{CoRR}, abs/2012.08452, 2020.

\bibitem[Bardes et~al.(2021)Bardes, Ponce, and LeCun]{vicreg}
Bardes, A., Ponce, J., and LeCun, Y.
\newblock Vicreg: Variance-invariance-covariance regularization for
  self-supervised learning.
\newblock \emph{CoRR}, abs/2105.04906, 2021.

\bibitem[Belghazi et~al.(2018)Belghazi, Baratin, Rajeswar, Ozair, Bengio,
  Hjelm, and Courville]{MINE}
Belghazi, M.~I., Baratin, A., Rajeswar, S., Ozair, S., Bengio, Y., Hjelm,
  R.~D., and Courville, A.~C.
\newblock Mutual information neural estimation.
\newblock In \emph{Proceedings of the 35th International Conference on Machine
  Learning, {ICML} 2018, Stockholm, Sweden, July 10-15, 2018}, volume~80 of
  \emph{Proceedings of Machine Learning Research}, pp.\  530--539, 2018.

\bibitem[Bemis \& Murcko(1996)Bemis and Murcko]{ScaffoldSplit}
Bemis, G.~W. and Murcko, M.~A.
\newblock The properties of known drugs. 1. molecular frameworks.
\newblock \emph{Journal of Medicinal Chemistry}, 39\penalty0 (15):\penalty0
  2887--2893, 1996.

\bibitem[Brockschmidt(2020)]{GNNFilm}
Brockschmidt, M.
\newblock Gnn-film: Graph neural networks with feature-wise linear modulation.
\newblock In \emph{Proceedings of the 37th International Conference on Machine
  Learning, {ICML} 2020, 13-18 July 2020, Virtual Event}, volume 119 of
  \emph{Proceedings of Machine Learning Research}, pp.\  1144--1152. {PMLR},
  2020.

\bibitem[Bronstein et~al.(2021)Bronstein, Bruna, Cohen, and Velickovic]{5G}
Bronstein, M.~M., Bruna, J., Cohen, T., and Velickovic, P.
\newblock Geometric deep learning: Grids, groups, graphs, geodesics, and
  gauges.
\newblock \emph{CoRR}, abs/2104.13478, 2021.

\bibitem[Caron et~al.(2020)Caron, Misra, Mairal, Goyal, Bojanowski, and
  Joulin]{SwAV}
Caron, M., Misra, I., Mairal, J., Goyal, P., Bojanowski, P., and Joulin, A.
\newblock Unsupervised learning of visual features by contrasting cluster
  assignments.
\newblock In Larochelle, H., Ranzato, M., Hadsell, R., Balcan, M., and Lin, H.
  (eds.), \emph{Advances in Neural Information Processing Systems 33: Annual
  Conference on Neural Information Processing Systems 2020, NeurIPS 2020,
  December 6-12, 2020, virtual}, 2020.

\bibitem[Chen et~al.(2020{\natexlab{a}})Chen, Liu, Hsieh, Chen, and
  Heng]{chen2020utilizing}
Chen, P., Liu, W., Hsieh, C.-Y., Chen, G., and Heng, P.~A.
\newblock Utilizing edge features in graph neural networks via variational
  information maximization, 2020{\natexlab{a}}.

\bibitem[Chen et~al.(2020{\natexlab{b}})Chen, Kornblith, Norouzi, and
  Hinton]{SimCLR}
Chen, T., Kornblith, S., Norouzi, M., and Hinton, G.~E.
\newblock A simple framework for contrastive learning of visual
  representations.
\newblock In \emph{Proceedings of the 37th International Conference on Machine
  Learning, {ICML} 2020, 13-18 July 2020, Virtual Event}, volume 119 of
  \emph{Proceedings of Machine Learning Research}, pp.\  1597--1607. {PMLR},
  2020{\natexlab{b}}.

\bibitem[Chen \& He(2020)Chen and He]{SimSiam}
Chen, X. and He, K.
\newblock Exploring simple siamese representation learning.
\newblock \emph{CoRR}, abs/2011.10566, 2020.

\bibitem[Coley et~al.(2019)Coley, Jin, Rogers, Jamison, Jaakkola, Green,
  Barzilay, and Jensen]{property_predGCNBarzilay}
Coley, C., Jin, W., Rogers, L., Jamison, T.~F., Jaakkola, T.~S., Green, W.~H.,
  Barzilay, R., and Jensen, K.~F.
\newblock A graph-convolutional neural network model for the prediction of
  chemical reactivity.
\newblock \emph{Chem. Sci.}, 10:\penalty0 370--377, 2019.

\bibitem[Corso et~al.(2020)Corso, Cavalleri, Beaini, Li\`{o}, and
  Veli\v{c}kovi\'{c}]{pna}
Corso, G., Cavalleri, L., Beaini, D., Li\`{o}, P., and Veli\v{c}kovi\'{c}, P.
\newblock Principal neighbourhood aggregation for graph nets.
\newblock In Larochelle, H., Ranzato, M., Hadsell, R., Balcan, M.~F., and Lin,
  H. (eds.), \emph{Advances in Neural Information Processing Systems},
  volume~33, pp.\  13260--13271. Curran Associates, Inc., 2020.

\bibitem[Dral(2020)]{quantum_chemistry}
Dral, P.~O.
\newblock Quantum chemistry in the age of machine learning.
\newblock \emph{The Journal of Physical Chemistry Letters}, 11\penalty0
  (6):\penalty0 2336--2347, 2020.

\bibitem[Fey \& Lenssen(2019)Fey and Lenssen]{Fey/Lenssen/2019}
Fey, M. and Lenssen, J.~E.
\newblock Fast graph representation learning with {PyTorch Geometric}.
\newblock In \emph{ICLR Workshop on Representation Learning on Graphs and
  Manifolds}, 2019.

\bibitem[Ganea et~al.(2021)Ganea, Pattanaik, Coley, Barzilay, Jensen, Green,
  and Jaakkola]{ganea2021geomol}
Ganea, O.-E., Pattanaik, L., Coley, C.~W., Barzilay, R., Jensen, K.~F., Green,
  W.~H., and Jaakkola, T.~S.
\newblock Geomol: Torsional geometric generation of molecular 3d conformer
  ensembles, 2021.

\bibitem[Gilmer et~al.(2017)Gilmer, Schoenholz, Riley, Vinyals, and Dahl]{mpnn}
Gilmer, J., Schoenholz, S.~S., Riley, P.~F., Vinyals, O., and Dahl, G.~E.
\newblock Neural message passing for quantum chemistry.
\newblock In Teh, Y.~W. (ed.), \emph{Proceedings of the 34th International
  Conference on Machine Learning, {ICML} 2017, Sydney, NSW, Australia, 6-11
  August 2017}, volume~70 of \emph{Proceedings of Machine Learning Research},
  pp.\  1263--1272. {PMLR}, 2017.

\bibitem[Gorgulla et~al.(2020)Gorgulla, Boeszoermenyi, Wang, Fischer, Coote,
  Padmanabha~Das, Malets, Radchenko, Moroz, Scott, Fackeldey, Hoffmann,
  Iavniuk, Wagner, and Arthanari]{enamine_real_database}
Gorgulla, C., Boeszoermenyi, A., Wang, Z.-F., Fischer, P.~D., Coote, P.~W.,
  Padmanabha~Das, K.~M., Malets, Y.~S., Radchenko, D.~S., Moroz, Y.~S., Scott,
  D.~A., Fackeldey, K., Hoffmann, M., Iavniuk, I., Wagner, G., and Arthanari,
  H.
\newblock An open-source drug discovery platform enables ultra-large virtual
  screens.
\newblock \emph{Nature}, 580\penalty0 (7805):\penalty0 663--668, 2020.

\bibitem[Gretton et~al.(2012)Gretton, Borgwardt, Rasch, Sch{{\"o}}lkopf, and
  Smola]{JMLR:v13:gretton12a}
Gretton, A., Borgwardt, K.~M., Rasch, M.~J., Sch{{\"o}}lkopf, B., and Smola, A.
\newblock A kernel two-sample test.
\newblock \emph{Journal of Machine Learning Research}, 13\penalty0
  (25):\penalty0 723--773, 2012.

\bibitem[Grill et~al.(2020)Grill, Strub, Altch{\'{e}}, Tallec, Richemond,
  Buchatskaya, Doersch, Pires, Guo, Azar, Piot, Kavukcuoglu, Munos, and
  Valko]{BYOL}
Grill, J., Strub, F., Altch{\'{e}}, F., Tallec, C., Richemond, P.~H.,
  Buchatskaya, E., Doersch, C., Pires, B.~{\'{A}}., Guo, Z., Azar, M.~G., Piot,
  B., Kavukcuoglu, K., Munos, R., and Valko, M.
\newblock Bootstrap your own latent - {A} new approach to self-supervised
  learning.
\newblock In Larochelle, H., Ranzato, M., Hadsell, R., Balcan, M., and Lin, H.
  (eds.), \emph{Advances in Neural Information Processing Systems 33: Annual
  Conference on Neural Information Processing Systems 2020, NeurIPS 2020,
  December 6-12, 2020, virtual}, 2020.

\bibitem[Grimme(2019)]{crest}
Grimme, S.
\newblock Exploration of chemical compound, conformer, and reaction space with
  meta-dynamics simulations based on tight-binding quantum chemical
  calculations.
\newblock \emph{Journal of Chemical Theory and Computation}, 15\penalty0
  (5):\penalty0 2847--2862, 2019.

\bibitem[Gutmann \& Hyv{\"{a}}rinen(2010)Gutmann and Hyv{\"{a}}rinen]{NCE}
Gutmann, M. and Hyv{\"{a}}rinen, A.
\newblock Noise-contrastive estimation: {A} new estimation principle for
  unnormalized statistical models.
\newblock In Teh, Y.~W. and Titterington, D.~M. (eds.), \emph{Proceedings of
  the Thirteenth International Conference on Artificial Intelligence and
  Statistics, {AISTATS} 2010, Chia Laguna Resort, Sardinia, Italy, May 13-15,
  2010}, volume~9 of \emph{{JMLR} Proceedings}, pp.\  297--304. JMLR.org, 2010.

\bibitem[Hjelm et~al.(2019)Hjelm, Fedorov, Lavoie{-}Marchildon, Grewal,
  Bachman, Trischler, and Bengio]{DIM}
Hjelm, R.~D., Fedorov, A., Lavoie{-}Marchildon, S., Grewal, K., Bachman, P.,
  Trischler, A., and Bengio, Y.
\newblock Learning deep representations by mutual information estimation and
  maximization.
\newblock In \emph{7th International Conference on Learning Representations,
  {ICLR} 2019, New Orleans, LA, USA, May 6-9, 2019}. OpenReview.net, 2019.

\bibitem[Hu et~al.(2020{\natexlab{a}})Hu, Fey, Zitnik, Dong, Ren, Liu, Catasta,
  and Leskovec]{ogb2020hu}
Hu, W., Fey, M., Zitnik, M., Dong, Y., Ren, H., Liu, B., Catasta, M., and
  Leskovec, J.
\newblock Open graph benchmark: Datasets for machine learning on graphs.
\newblock In Larochelle, H., Ranzato, M., Hadsell, R., Balcan, M., and Lin, H.
  (eds.), \emph{Advances in Neural Information Processing Systems 33: Annual
  Conference on Neural Information Processing Systems 2020, NeurIPS 2020,
  December 6-12, 2020, virtual}, 2020{\natexlab{a}}.

\bibitem[Hu et~al.(2020{\natexlab{b}})Hu, Liu, Gomes, Zitnik, Liang, Pande, and
  Leskovec]{strategies_for_pretraining_gnn_hu_2020}
Hu, W., Liu, B., Gomes, J., Zitnik, M., Liang, P., Pande, V.~S., and Leskovec,
  J.
\newblock Strategies for pre-training graph neural networks.
\newblock In \emph{8th International Conference on Learning Representations,
  {ICLR} 2020, Addis Ababa, Ethiopia, April 26-30, 2020}. OpenReview.net,
  2020{\natexlab{b}}.

\bibitem[Hy et~al.(2018)Hy, Trivedi, Pan, Anderson, and
  Kondor]{molecularPropPredCompositionalNets}
Hy, T.~S., Trivedi, S., Pan, H., Anderson, B.~M., and Kondor, R.
\newblock Predicting molecular properties with covariant compositional
  networks.
\newblock \emph{The Journal of Chemical Physics}, 148\penalty0 (24):\penalty0
  241745, 2018.

\bibitem[Isert et~al.(2021)Isert, Atz, Jiménez-Luna, and
  Schneider]{isert2021qmugs}
Isert, C., Atz, K., Jiménez-Luna, J., and Schneider, G.
\newblock Qmugs: Quantum mechanical properties of drug-like molecules, 2021.

\bibitem[Jumper et~al.(2021)Jumper, Evans, Pritzel, Green, Figurnov,
  Ronneberger, Tunyasuvunakool, Bates, {\v{Z}}{\'i}dek, Potapenko, Bridgland,
  Meyer, Kohl, Ballard, Cowie, Romera-Paredes, Nikolov, Jain, Adler, Back,
  Petersen, Reiman, Clancy, Zielinski, Steinegger, Pacholska, Berghammer,
  Bodenstein, Silver, Vinyals, Senior, Kavukcuoglu, Kohli, and
  Hassabis]{alphafold2}
Jumper, J., Evans, R., Pritzel, A., Green, T., Figurnov, M., Ronneberger, O.,
  Tunyasuvunakool, K., Bates, R., {\v{Z}}{\'i}dek, A., Potapenko, A.,
  Bridgland, A., Meyer, C., Kohl, S. A.~A., Ballard, A.~J., Cowie, A.,
  Romera-Paredes, B., Nikolov, S., Jain, R., Adler, J., Back, T., Petersen, S.,
  Reiman, D., Clancy, E., Zielinski, M., Steinegger, M., Pacholska, M.,
  Berghammer, T., Bodenstein, S., Silver, D., Vinyals, O., Senior, A.~W.,
  Kavukcuoglu, K., Kohli, P., and Hassabis, D.
\newblock Highly accurate protein structure prediction with alphafold.
\newblock \emph{Nature}, 596\penalty0 (7873):\penalty0 583--589, 8 2021.

\bibitem[Klicpera et~al.(2020{\natexlab{a}})Klicpera, Giri, Margraf, and
  G{\"u}nnemann]{klicpera_dimenetpp_2020}
Klicpera, J., Giri, S., Margraf, J.~T., and G{\"u}nnemann, S.
\newblock Fast and uncertainty-aware directional message passing for
  non-equilibrium molecules.
\newblock In \emph{NeurIPS-W}, 2020{\natexlab{a}}.

\bibitem[Klicpera et~al.(2020{\natexlab{b}})Klicpera, Gro{\ss}, and
  G{\"u}nnemann]{klicpera_dimenet_2020}
Klicpera, J., Gro{\ss}, J., and G{\"u}nnemann, S.
\newblock Directional message passing for molecular graphs.
\newblock In \emph{International Conference on Learning Representations
  (ICLR)}, 2020{\natexlab{b}}.

\bibitem[Klicpera et~al.(2021)Klicpera, Becker, and
  Günnemann]{klicpera2021gemnet}
Klicpera, J., Becker, F., and Günnemann, S.
\newblock Gemnet: Universal directional graph neural networks for molecules.
\newblock \emph{NeurIPS}, 2021.

\bibitem[Landrum(2016)]{rdkit}
Landrum, G.
\newblock Rdkit: Open-source cheminformatics software, 2016.

\bibitem[Li et~al.(2017)Li, Cai, and He]{GRLforDrugDIscorver}
Li, J., Cai, D., and He, X.
\newblock Learning graph-level representation for drug discovery.
\newblock \emph{CoRR}, abs/1709.03741, 2017.

\bibitem[Liu et~al.(2022)Liu, Wang, Liu, Lasenby, Guo, and
  Tang]{liu2022pretraining}
Liu, S., Wang, H., Liu, W., Lasenby, J., Guo, H., and Tang, J.
\newblock Pre-training molecular graph representation with 3d geometry.
\newblock \emph{10th International Conference on Learning Representations,
  {ICLR} 2022}, 2022.

\bibitem[Liu et~al.(2021)Liu, Wang, Liu, Zhang, Oztekin, and
  Ji]{sphericalmessagepassing2021}
Liu, Y., Wang, L., Liu, M., Zhang, X., Oztekin, B., and Ji, S.
\newblock Spherical message passing for 3d graph networks.
\newblock \emph{CoRR}, abs/2102.05013, 2021.

\bibitem[Pan \& Yang(2010)Pan and Yang]{negative_transfer}
Pan, S.~J. and Yang, Q.
\newblock A survey on transfer learning.
\newblock \emph{{IEEE} Trans. Knowl. Data Eng.}, 22\penalty0 (10):\penalty0
  1345--1359, 2010.

\bibitem[Paszke et~al.(2017)Paszke, Gross, Chintala, Chanan, Yang, DeVito, Lin,
  Desmaison, Antiga, and Lerer]{paszke2017automatic}
Paszke, A., Gross, S., Chintala, S., Chanan, G., Yang, E., DeVito, Z., Lin, Z.,
  Desmaison, A., Antiga, L., and Lerer, A.
\newblock Automatic differentiation in pytorch, 2017.

\bibitem[Rahaman et~al.(2019)Rahaman, Baratin, Arpit, Draxler, Lin, Hamprecht,
  Bengio, and Courville]{spectral_bias_of_nn}
Rahaman, N., Baratin, A., Arpit, D., Draxler, F., Lin, M., Hamprecht, F.,
  Bengio, Y., and Courville, A.
\newblock On the spectral bias of neural networks.
\newblock In Chaudhuri, K. and Salakhutdinov, R. (eds.), \emph{Proceedings of
  the 36th International Conference on Machine Learning}, volume~97 of
  \emph{Proceedings of Machine Learning Research}, pp.\  5301--5310. PMLR, 6
  2019.

\bibitem[Ramakrishnan et~al.(2014)Ramakrishnan, Dral, Rupp, and von
  Lilienfeld]{Ramakrishnan2014}
Ramakrishnan, R., Dral, P.~O., Rupp, M., and von Lilienfeld, O.~A.
\newblock Quantum chemistry structures and properties of 134 kilo molecules.
\newblock \emph{Scientific Data}, 1\penalty0 (1):\penalty0 140022, 8 2014.

\bibitem[Satorras et~al.(2021)Satorras, Hoogeboom, and Welling]{egnn}
Satorras, V.~G., Hoogeboom, E., and Welling, M.
\newblock E(n) equivariant graph neural networks.
\newblock \emph{CoRR}, abs/2102.09844, 2021.

\bibitem[Schmidt et~al.(2019)Schmidt, Marques, Botti, and Marques]{Schmidt2019}
Schmidt, J., Marques, M. R.~G., Botti, S., and Marques, M. A.~L.
\newblock Recent advances and applications of machine learning in solid-state
  materials science.
\newblock \emph{npj Computational Materials}, 5\penalty0 (1):\penalty0 83, 8
  2019.

\bibitem[Sch{\"{u}}tt et~al.(2017)Sch{\"{u}}tt, Kindermans, Felix, Chmiela,
  Tkatchenko, and M{\"{u}}ller]{SchNet}
Sch{\"{u}}tt, K., Kindermans, P., Felix, H. E.~S., Chmiela, S., Tkatchenko, A.,
  and M{\"{u}}ller, K.
\newblock Schnet: {A} continuous-filter convolutional neural network for
  modeling quantum interactions.
\newblock In Guyon, I., von Luxburg, U., Bengio, S., Wallach, H.~M., Fergus,
  R., Vishwanathan, S. V.~N., and Garnett, R. (eds.), \emph{Advances in Neural
  Information Processing Systems 30: Annual Conference on Neural Information
  Processing Systems 2017, December 4-9, 2017, Long Beach, CA, {USA}}, pp.\
  991--1001, 2017.

\bibitem[Shi et~al.(2021)Shi, Luo, Xu, and Tang]{LearningGradFields}
Shi, C., Luo, S., Xu, M., and Tang, J.
\newblock Learning gradient fields for molecular conformation generation.
\newblock In Meila, M. and Zhang, T. (eds.), \emph{Proceedings of the 38th
  International Conference on Machine Learning, {ICML} 2021, 18-24 July 2021,
  Virtual Event}, volume 139 of \emph{Proceedings of Machine Learning
  Research}, pp.\  9558--9568. {PMLR}, 2021.

\bibitem[Stokes et~al.(2020)Stokes, Yang, Swanson, Jin, Cubillos-Ruiz, Donghia,
  MacNair, French, Carfrae, Bloom-Ackermann, Tran, Chiappino-Pepe, Badran,
  Andrews, Chory, Church, Brown, Jaakkola, Barzilay, and Collins]{antibiotic}
Stokes, J.~M., Yang, K., Swanson, K., Jin, W., Cubillos-Ruiz, A., Donghia,
  N.~M., MacNair, C.~R., French, S., Carfrae, L.~A., Bloom-Ackermann, Z., Tran,
  V.~M., Chiappino-Pepe, A., Badran, A.~H., Andrews, I.~W., Chory, E.~J.,
  Church, G.~M., Brown, E.~D., Jaakkola, T.~S., Barzilay, R., and Collins,
  J.~J.
\newblock A deep learning approach to antibiotic discovery.
\newblock \emph{Cell}, 180\penalty0 (4):\penalty0 688--702.e13, 2020.

\bibitem[Tancik et~al.(2020)Tancik, Srinivasan, Mildenhall, Fridovich-Keil,
  Raghavan, Singhal, Ramamoorthi, Barron, and Ng]{tancik2020fourfeat}
Tancik, M., Srinivasan, P.~P., Mildenhall, B., Fridovich-Keil, S., Raghavan,
  N., Singhal, U., Ramamoorthi, R., Barron, J.~T., and Ng, R.
\newblock Fourier features let networks learn high frequency functions in low
  dimensional domains.
\newblock \emph{NeurIPS}, 2020.

\bibitem[Tang et~al.(2020)Tang, Kramer, Fang, Qiu, Wu, and Xu]{Tang}
Tang, B., Kramer, S.~T., Fang, M., Qiu, Y., Wu, Z., and Xu, D.
\newblock A self-attention based message passing neural network for predicting
  molecular lipophilicity and aqueous solubility.
\newblock \emph{J. Cheminformatics}, 12\penalty0 (1):\penalty0 15, 2020.

\bibitem[Tian et~al.(2019)Tian, Krishnan, and Isola]{distillation2019Tian}
Tian, Y., Krishnan, D., and Isola, P.
\newblock Contrastive representation distillation.
\newblock \emph{CoRR}, abs/1910.10699, 2019.
\newblock URL \url{http://arxiv.org/abs/1910.10699}.

\bibitem[Torng \& Altman(2019)Torng and Altman]{drug_target_GCN}
Torng, W. and Altman, R.~B.
\newblock Graph convolutional neural networks for predicting drug-target
  interactions.
\newblock \emph{J. Chem. Inf. Model.}, 59\penalty0 (10):\penalty0 4131--4149,
  2019.

\bibitem[Unke \& Meuwly(2019)Unke and Meuwly]{PhysNet}
Unke, O.~T. and Meuwly, M.
\newblock Physnet: A neural network for predicting energies, forces, dipole
  moments, and partial charges.
\newblock \emph{Journal of Chemical Theory and Computation}, 15\penalty0
  (6):\penalty0 3678--3693, 2019.

\bibitem[van~den Oord et~al.(2018)van~den Oord, Li, and Vinyals]{CPC}
van~den Oord, A., Li, Y., and Vinyals, O.
\newblock Representation learning with contrastive predictive coding.
\newblock \emph{CoRR}, abs/1807.03748, 2018.

\bibitem[Wang et~al.(2019)Wang, Zheng, Ye, Gan, Li, Song, Zhou, Ma, Yu, Gai,
  Xiao, He, Karypis, Li, and Zhang]{wang2019dgl}
Wang, M., Zheng, D., Ye, Z., Gan, Q., Li, M., Song, X., Zhou, J., Ma, C., Yu,
  L., Gai, Y., Xiao, T., He, T., Karypis, G., Li, J., and Zhang, Z.
\newblock Deep graph library: A graph-centric, highly-performant package for
  graph neural networks.
\newblock \emph{arXiv preprint arXiv:1909.01315}, 2019.

\bibitem[Wang et~al.(2021)Wang, Wang, Cao, and Farimani]{MolCLR}
Wang, Y., Wang, J., Cao, Z., and Farimani, A.~B.
\newblock Molclr: Molecular contrastive learning of representations via graph
  neural networks.
\newblock \emph{CoRR}, abs/2102.10056, 2021.

\bibitem[Withnall et~al.(2020)Withnall, Lindel{\"{o}}f, Engkvist, and
  Chen]{mpnnForPysicalChemical}
Withnall, M., Lindel{\"{o}}f, E., Engkvist, O., and Chen, H.
\newblock Building attention and edge message passing neural networks for
  bioactivity and physical-chemical property prediction.
\newblock \emph{J. Cheminformatics}, 12\penalty0 (1):\penalty0 1, 2020.

\bibitem[Wu et~al.(2017)Wu, Ramsundar, Feinberg, Gomes, Geniesse, Pappu,
  Leswing, and Pande]{moleculenet}
Wu, Z., Ramsundar, B., Feinberg, E.~N., Gomes, J., Geniesse, C., Pappu, A.~S.,
  Leswing, K., and Pande, V.~S.
\newblock Moleculenet: {A} benchmark for molecular machine learning.
\newblock \emph{CoRR}, abs/1703.00564, 2017.

\bibitem[Xu et~al.(2021{\natexlab{a}})Xu, Luo, Bengio, Peng, and
  Tang]{LearningGenerativeDyn_MinkaiXu}
Xu, M., Luo, S., Bengio, Y., Peng, J., and Tang, J.
\newblock Learning neural generative dynamics for molecular conformation
  generation.
\newblock In \emph{9th International Conference on Learning Representations,
  {ICLR} 2021, Virtual Event, Austria, May 3-7, 2021}. OpenReview.net,
  2021{\natexlab{a}}.

\bibitem[Xu et~al.(2021{\natexlab{b}})Xu, Wang, Ni, Guo, and Tang]{minghao}
Xu, M., Wang, H., Ni, B., Guo, H., and Tang, J.
\newblock Self-supervised graph-level representation learning with local and
  global structure.
\newblock In Meila, M. and Zhang, T. (eds.), \emph{Proceedings of the 38th
  International Conference on Machine Learning, {ICML} 2021, 18-24 July 2021,
  Virtual Event}, volume 139 of \emph{Proceedings of Machine Learning
  Research}, pp.\  11548--11558. {PMLR}, 2021{\natexlab{b}}.

\bibitem[Yang et~al.(2019)Yang, Swanson, Jin, Coley, Eiden, Gao, Guzman-Perez,
  Hopper, Kelley, Mathea, Palmer, Settels, Jaakkola, Jensen, and
  Barzilay]{chemprop}
Yang, K., Swanson, K., Jin, W., Coley, C., Eiden, P., Gao, H., Guzman-Perez,
  A., Hopper, T., Kelley, B., Mathea, M., Palmer, A., Settels, V., Jaakkola,
  T., Jensen, K., and Barzilay, R.
\newblock Analyzing learned molecular representations for property prediction.
\newblock \emph{Journal of Chemical Information and Modeling}, 59\penalty0
  (8):\penalty0 3370--3388, 2019.

\bibitem[You et~al.(2020)You, Chen, Sui, Chen, Wang, and Shen]{GraphCL}
You, Y., Chen, T., Sui, Y., Chen, T., Wang, Z., and Shen, Y.
\newblock Graph contrastive learning with augmentations.
\newblock In Larochelle, H., Ranzato, M., Hadsell, R., Balcan, M., and Lin, H.
  (eds.), \emph{Advances in Neural Information Processing Systems 33: Annual
  Conference on Neural Information Processing Systems 2020, NeurIPS 2020,
  December 6-12, 2020, virtual}, 2020.

\bibitem[You et~al.(2021)You, Chen, Shen, and Wang]{GraphCLautomated}
You, Y., Chen, T., Shen, Y., and Wang, Z.
\newblock Graph contrastive learning automated.
\newblock In Meila, M. and Zhang, T. (eds.), \emph{Proceedings of the 38th
  International Conference on Machine Learning, {ICML} 2021, 18-24 July 2021,
  Virtual Event}, volume 139 of \emph{Proceedings of Machine Learning
  Research}, pp.\  12121--12132. {PMLR}, 2021.

\bibitem[Zbontar et~al.(2021)Zbontar, Jing, Misra, LeCun, and
  Deny]{BarlowTwins}
Zbontar, J., Jing, L., Misra, I., LeCun, Y., and Deny, S.
\newblock Barlow twins: Self-supervised learning via redundancy reduction.
\newblock In Meila, M. and Zhang, T. (eds.), \emph{Proceedings of the 38th
  International Conference on Machine Learning, {ICML} 2021, 18-24 July 2021,
  Virtual Event}, volume 139 of \emph{Proceedings of Machine Learning
  Research}, pp.\  12310--12320. {PMLR}, 2021.

\bibitem[Zhu et~al.(2021)Zhu, Xia, Qin, Zhou, Li, and Liu]{ZhuDualView}
Zhu, J., Xia, Y., Qin, T., Zhou, W., Li, H., and Liu, T.
\newblock Dual-view molecule pre-training.
\newblock \emph{CoRR}, abs/2106.10234, 2021.

\end{thebibliography}
\bibliographystyle{icml2022}

\newpage
\appendix
\onecolumn

\section{Further Explanations}\label{Net3d-details}
\textbf{3D Network Details}
The l-th layer of the 3D network takes two sets as input. First, $n^2-n$ edge representations $\{d^l_{uv} \in \mathbb{R}^{d_d} \mid u,v \in \mathcal{V} \land u\neq v\}$ (the edges of a complete graph without self-loops). In the first layer they are given by the encoded distances fed through an initial feed-forward network  $U_{init}: \mathbb{R}^{2F+1} \mapsto \mathbb{R}^{d_d}$ which projects them to the hidden dimension of the edges $d^0_{uv}= U_{init}(\gamma(d_{uv}))$. The second input is a set of $n$ atom representations $\{h^l_1, \dots h^l_n\}$ with dimensionality $\mathbb{R}^{d_h}$. In the first layer, the atom representations are all set to the same learned vector that is initialized with a standard normal.
With $\mathrel{\Vert}$ meaning concatenation, every layer updates the edge and atom representations and iteratively encodes 3D information into them as follows:
\begin{equation}
    m_{uv} = U_{edge}([h^l_u \mathrel{\Vert} h^l_v \mathrel{\Vert}d_{uv}^l])
\end{equation}
\begin{equation}
    d_{uv}^{l +1} = d_{uv}^l + m_{uv}
\end{equation}
\begin{equation}
    h_{u}^{l +1} = U_{h}( [h_{u} \mathrel{\Vert} \sum_{\substack{v = 1 \\ v\neq u}}^n  m_{uv}* \sigma(U_{softedge}(m_{uv})]).
\end{equation}
The layer is parameterized by three MLPs where the first one updates the edges $U_{edge}: \mathbb{R}^{2d_h + d_d} \mapsto \mathbb{R}^{d_d}$. The second one updates the atom representations $U_{h}: \mathbb{R}^{d_h + d_d} \mapsto \mathbb{R}^{d_h}$. The third one $U_{softedge}: \mathbb{R}^{d_d} \mapsto \mathbb{R}$ is followed by the logistic sigmoid function to create a value between 0 and 1 that can be seen as a soft edge weight telling us how probable an edge is for each message $m_{uv}$ as it is done by \citet{egnn}.

To produce the final 3D representation $z^b$, all atom representations are aggregated by concatenating their mean, their maximum, and their standard deviation and feeding this through a final feed-forward network $U: \mathbb{R}^{3d_h} \mapsto \mathbb{R}^{d_z}$.

\begin{figure}[htpb]
  \centering
  \includegraphics[width=\textwidth]{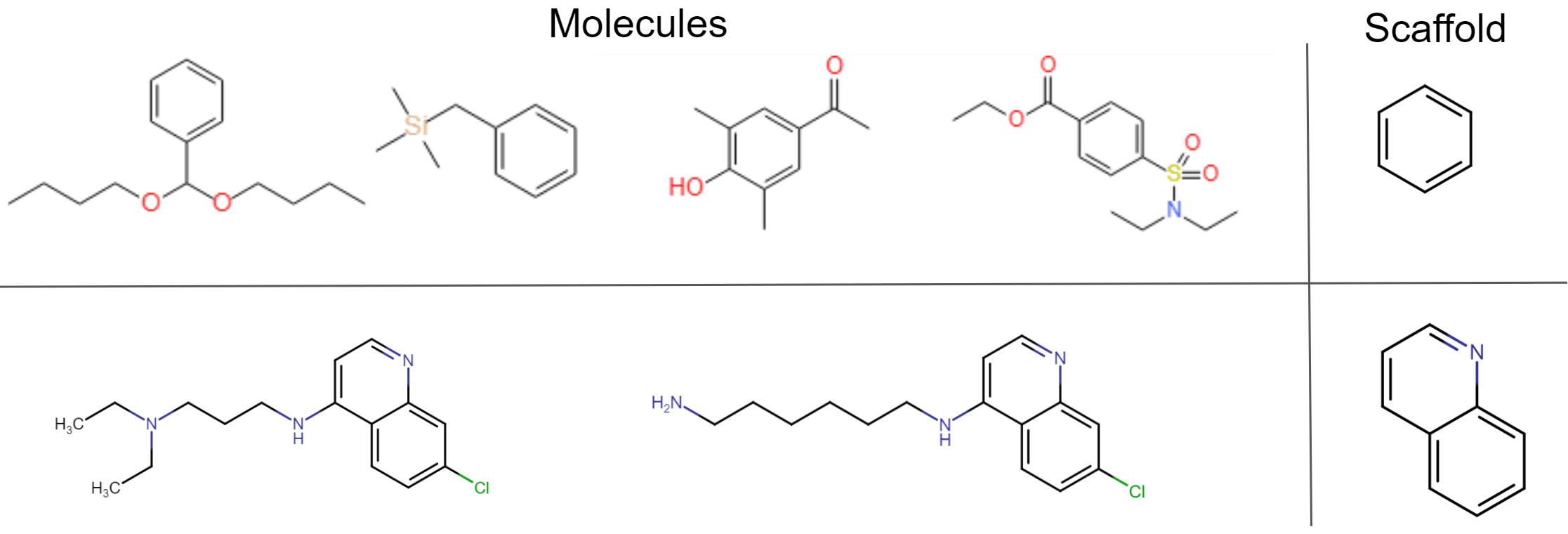}
  \caption[Depiction of two groups of molecules and their Bemis-Murcko scaffold to explain the concept of scaffolds.]{Depiction of two groups of molecules where all the molecules in the top row have the same Bemis-Murcko scaffold \cite{ScaffoldSplit} which is different from the scaffold to which the two molecules in the bottom row belong. We can easily obtain the scaffold of a molecule using RDKit \cite{rdkit}.}\label{scaffold-image}
\end{figure}

\section{Experimental Details}\label{experimental-details-appendix}

\subsection{Parameter Details}\label{parameter-details}
The hyperparameters for SMP are taken from the official repository\footnote{\url{https://github.com/divelab/DIG}} where \citet{sphericalmessagepassing2021} provide their code, and we predict $gap$ even though it could be calculated as $|homo - lumo|$.
The parameter search space and final parameters for the PNA architecture are specified in Table \ref{pna-details-table} and those of the 3D network in Table \ref{net-3d-details-table}. 

\textbf{Pre-training:} We use Adam with a start learning rate of $8\times 10^{-5}$ and a batch size of 500. The learning rate schedule during pre-training starts with 700 optimization steps of linear warmup followed by the schedule given by the \textit{ReduceLROnPlateau} scheduler by PyTorch\footnote{\url{https://pytorch.org/docs/stable/generated/torch.optim.lr_scheduler.ReduceLROnPlateau.html}} with reduction parameter 0.6, patience 25, and a cooldown of 20.

\textbf{Fine-tuning quantum mechanical properties:} We use Adam with a start learning rate of $7\times 10^{-5}$, weight decay $1\times 10^{-11}$ and a batch size of 128. For the learning rate schedule, we first perform warmup as follows. We consider three different sets of learnable parameters: (1) the batch norm parameters, (2) all newly initialized parameters that were not transferred, and (3) all parameters. For these sets, we increase the learning rate in this order from 0 to the start learning rate with linear interpolation. For parameter group one, we warm up for 700 steps, 700 steps for group 2, and 350 steps for group 3. After that 
we use the schedule given by the \textit{ReduceLROnPlateau} with reduction parameter 0.5, patience 25, and a cooldown of 20.

\textbf{Fine-tuning non-quantum properties:} We use Adam with a start learning rate of $1\times 10^{-3}$ and a batch size of 32. The learning rate schedule is the same as for the quantum mechanical properties.

The experiment on the non-quantum properties has different hyperparameters for PNA since the smaller datasets are easily overfitted on with the large architecture we use for the quantum mechanical properties. A smaller PNA yields better performance with the random initialization baseline. Therefore the PNA in these experiments has a hidden dimension of 50 and 3 message passing layers as propagation depth. Apart from that, it is the same as the PNA described in Table \ref{pna-details-table}.

\begin{table}[htpb]
\caption[Search space for the 2D network PNA.]{Search space for the 2D network PNA through which we searched to obtain a strong baseline performance on the homo property of the  QM9 dataset. The parameters were tuned in the order in which they are listed in this table from top to bottom. After this was completed for all parameters, we performed a second round of tuning for a subset of them. The final parameters are marked in \textbf{bold}.}
\label{pna-details-table}
\begin{center}
\begin{small}
\begin{sc}
\begin{tabular}{lc}
\toprule
Parameter & Search Space  \\    
\midrule
propagation depth & [4, 5, 6 ,\textbf{7}] \\
hidden dimension &  [40, 50, 75, 90, 100, 150, \textbf{200} ,300]  \\
message MLP layers & [1, \textbf{2}, 3] \\
update MLP layers & [\textbf{1}, 2, 3] \\
aggregators & \makecell[c]{[mean, max, min, std, sum], [mean, max, min], [mean, max, sum], \\ \textbf{[mean, max, min, std]}, [max, sum], [sum] } \\
scalers & [identity],  \textbf{[identity, amplification, attenuation]} \\
readout aggregators & [mean], [sum], [mean, max, sum], \textbf{[mean, max, min, sum]}\\
dropout & [\textbf{0}, 0.05, 0.1, 0.2] \\
batchnorm after MLPs & \textbf{True}/False \\
batchnorm in MLPs & \textbf{True}/False\\
readout MLP layers & [1, \textbf{2}, 3] \\
batchnorm momentum & [\textbf{0.1}, 0.9, 0.93] \\
\bottomrule
\end{tabular}
\end{sc}
\end{small}
\end{center}
\vskip -0.1in
\end{table}

\begin{table}[htpb]
\caption{Search space for the 3D network \textit{Net3D} through which we searched to obtain a strong baseline performance on the homo property of the  QM9 dataset and we considered the size of the network where parameters leading to less memory use are preferred. The parameters were tuned in the order in which they are listed in this table from top to bottom. After this was completed for all parameters, we performed a second round of tuning for a subset of them. The final parameters are marked in \textbf{bold}.}
\label{net-3d-details-table}
\begin{center}
\begin{small}
\begin{sc}
\begin{tabular}{lc}
\toprule
Parameter & Search Space  \\    
\midrule
propagation depth & [\textbf{1}, 3, 4, 5] \\
hidden dimension &  [10,  \textbf{20}, 40, 60, 80, 100]  \\
F used in $\gamma: \mathbb{R} \mapsto \mathbb{R}^{2F+1}$ & [0, 3, \textbf{4}, 8, 10, 50] \\
message MLP layers & [\textbf{1}, 2, 3] \\
update MLP layers & [\textbf{1}, 2, 3] \\
readout aggregators & [mean], [sum], \textbf{[mean, max, min]}, [mean, max, min, sum] \\
dropout & [\textbf{0}, 0.05, 0.1, 0.2, 0.5] \\
batchnorm after MLPs & \textbf{True}/False \\
readout MLP layers & [\textbf{1}, 2, 3] \\
batchnorm momentum & [0.1, 0.9, \textbf{0.93}] \\
\bottomrule
\end{tabular}
\end{sc}
\end{small}
\end{center}
\vskip -0.1in
\end{table}

\subsection{Data Details}\label{data-details}
\begin{figure}[htb]
  \centering
  \includegraphics[width=0.5\textwidth]{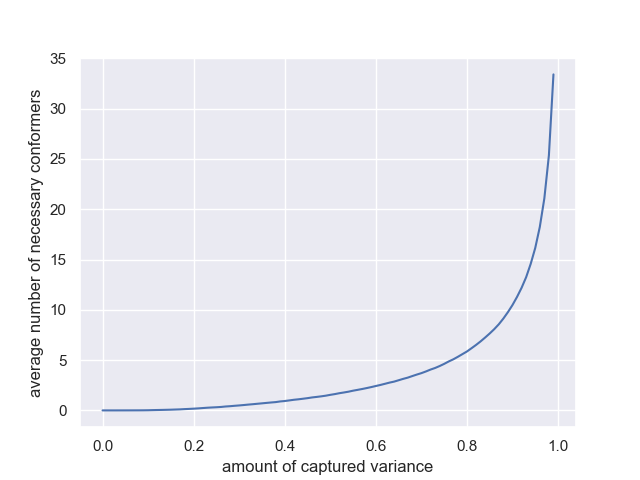}\label{conformers-per-boltzmannweight}
  \includegraphics[width=0.5\textwidth]{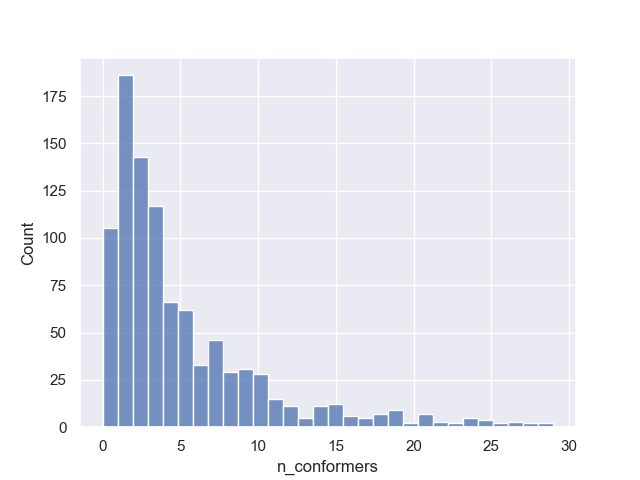}\label{conformers-drugs-histogram}
  \caption[]{a) The average number of conformers necessary to cover a certain amount of Boltzmann weight in GEOM-Drugs. For a given amount of cumulative Boltzmann weight on the horizontal axis, the vertical axis shows the average number of conformers necessary to pass that threshold. b) Histogram of how many molecules there are in GEOM-Drugs with a certain amount of conformers. The histogram is created for 1000 molecules of GEOM-Drugs.} \label{geom-drugs-boltzmannweight}
\end{figure}

We use three datasets containing 3D information for pre-training with diversity in molecule size and the number of molecules, as can be seen in Table \ref{dataset-table}. The pre-training datasets are:
\begin{enumerate}
    \item \textbf{QM9\footnote{\url{https://github.com/klicperajo/dimenet/blob/master/data/qm9_eV.npz}}} \cite{Ramakrishnan2014} contains 134k stable small organic molecules of 5 elements (CHONF). Every molecule has the 3D coordinates of one low-energy conformer and is annotated with 12 quantum mechanical properties as regression targets. The molecules are considered very small, with at most 9 heavy atoms.
    \item \textbf{GEOM-Drugs\footnote{\url{https://github.com/learningmatter-mit/geom}}} \cite{axelrod2020geom} consists of 304k realistically-sized biologically and pharmacologically relevant molecules of 16 elements, annotated with multiple 3D conformers, the ensemble Gibbs free energy, and the ensemble energy as regression targets. For the average molecule, 70\% of the Boltzmann weight is captured by just three conformers as can be seen in Figure \ref{conformers-per-boltzmannweight} where we also provide a histogram for the number of molecules that have a certain amount of conformers in Figure \ref{conformers-drugs-histogram}. The conformers are generated using CREST \cite{crest}.
    \item \textbf{QMugs\footnote{\url{https://www.research-collection.ethz.ch/handle/20.500.11850/482129}}} \cite{isert2021qmugs} has 665k drug-like molecules with three diverse conformers each and multiple conformer specific quantum mechanical properties as regression tasks. The conformers are generated using CREST \cite{crest}.
\end{enumerate}

For fine-tuning, we use a variety of datasets that cover a wide range of domains and applications. The molecular properties are relevant for quantum mechanics, physical chemistry, biophysics, and physiology such that we can obtain a good estimate of how valuable our 3D pre-training is for each domain. For quantum mechanical properties, which are often specific to a conformer, it is clear that 3D information is important and there has been a lot of evidence that learned methods highly benefit from its use \cite{klicpera_dimenet_2020, klicpera_dimenetpp_2020, sphericalmessagepassing2021,SchNet}. 
For these properties, the interest is in how much our method can leverage this information and transfer it to molecules where no 3D geometry is available.

Meanwhile, for biological or physiological properties such as blood-brain barrier penetration, it is not as clear if improvements from 3D information are to be expected. As such, this question needs to be answered next to how much of the benefits 3D pre-training recovers. For this purpose, we use the following molecular graph datasets, which are mainly taken from MoleculeNet \cite{moleculenet} and we use the scaffold splits\footnote{\url{https://ogb.stanford.edu/docs/graphprop}} with an 80/10/10 split ratio provided by OGB \citet{ogb2020hu}. The fine-tuning datasets are:

\begin{enumerate}[noitemsep]
    \item \textbf{QM9 and GEOM-Drugs:} On these 3D datasets we also fine-tune and evaluate the quantum mechanical properties of one half of the datasets with a random split. This is done after either pre-training on another 3D dataset (generalization), or after pre-training on the other half of the same dataset (in distribution).
    \item \textbf{ESOL:} $1128$ common organic small molecules with water solubility data (log solubility in mols per liter).
    \item \textbf{Lipo:} Experimental data for the octanol/water distribution coefficient of $4200$ molecules.
    \item \textbf{FreeSolv:} The hydration free energy of 642 molecules in water.
    \item \textbf{HIV:} 41k molecules with binary labels for HIV virus replication inhibition.
    \item \textbf{BACE:} Binary labels of binding results for inhibitors of human $\beta$-secretase 1 for 1512 molecules.
    \item \textbf{BBBP:} 2039 molecules with binary labels for blood-brain barrier penetration.
    \item \textbf{Tox21:} 7831 molecules with binary labels of their toxic for 12 different targets.
    \item \textbf{ToxCast:} 8576 molecules with binary labels of toxicity experiment outcomes with 617 targets.
    \item \textbf{SIDER:} 1427 approved drugs with 27 different side effect groups and the task is to predict whether the drug is in the side effect group.
    \item \textbf{ClinTox:} 1477 drugs with two binary annotations where the first is to predict toxicity in clinical trials and the second is the FDA approval status.
\end{enumerate}

The reason why \textit{muv} and \textit{pcba} are the only datasets from the OGB benchmark suite which we omit is their larger size.

\begin{table}[htb]
\caption[Statistics of the used datasets.]{Statistics of the used datasets. In the upper section are datasets with 3D information, which we use for pre-training, and the datasets in the bottom section do not contain additional 3D annotations.}
\label{dataset-table}
\begin{center}
\begin{small}
\begin{sc}
\begin{tabular}{lrrrc}
\toprule
Dataset   & \#Molecules & Avg.  \#Atoms & Avg. \#Bonds  & split   \\    
\midrule
\texttt{QM9          }                                 & 130\,831   & 18.0    & 18.6 & random  \\
\texttt{GEOM-Drugs   }                                 & 304\,293   & 44.4    & 46.4  & random \\
\texttt{QMugs}                                          &665\,911   & 30.6    & 33.4  & random \\
\hline
\texttt{esol         }                             & 1128 & 13.3 & 13.7  & scaffold\\
\texttt{lipo         }                                 &  4200 & 27.0 & 29.5 & scaffold\\
\texttt{freesolv     }                                     &  642 & 8.7 & 8.4  & scaffold\\
\texttt{bace         }                                  & 1512   & 34.1    & 36.9  & scaffold\\
\texttt{bbbp         }                                  & 2039   & 24.1    & 26.0  & scaffold\\
\texttt{hiv         }                                 &  41\,127 & 25.5 &  27.5 & scaffold \\
\texttt{tox21         }                                 &  7831 & 18.6 &  19.3 & scaffold \\
\texttt{toxcast         }                               & 8576  & 18.8 &  19.3 & scaffold \\
\texttt{clintox         }                                 &  1477 & 26.2 &  27.9 & scaffold \\
\texttt{sider         }                                 &  1427 & 33.6 &  35.4 & scaffold \\
\bottomrule
\end{tabular}
\end{sc}
\end{small}
\end{center}
\vskip -0.1in
\end{table}

\subsection{Implementation}\label{implementation}
Code to 3D pre-train a GNN or to reproduce results is available at 
\url{https://github.com/HannesStark/3DInfomax}.
All experiments were implemented in \textit{PyTorch} \cite{paszke2017automatic} using the deep learning libraries for processing graphs \textit{Pytorch Geometric} \cite{Fey/Lenssen/2019} and \textit{Deep Graph Library} \cite{wang2019dgl}. The code we use for SMP \cite{sphericalmessagepassing2021} is under the GNU General Public License v3.0 and we use their implementation after discussing it with the first author of the paper and under the consideration that their project welcomed our contributions to their library.

The experiments were conducted on two different machines while the same system was always used in direct comparisons. The first machine has an AMD Ryzen 1700 CPU @ 3.70Ghz, 16GB of RAM, and an Nvidia GTX 1060 GPU with 6GB vRAM.
The second system contains two Intel Xeon Gold 6248 CPUs @ 2.50GHz each with 20/40 cores, 400GB of RAM, and four Quadro RTX 8000 GPUs with 46GB vRAM of which only a single one was used for each experiment.
All mentions and of training time refer to the second system.

\subsection{Units and Meaning of Quantum Properties}\label{units-appendix}
For the GEOM-Drugs dataset, all reported numbers have the unit kcal/mol, Gibbs refers to the ensemble Gibbs free energy, and $\langle E \rangle$ to the ensemble energy.
\begin{table}[htpb]
\caption[Units of Quantum mechanical properties.]{Units and description of quantum mechanical properties of the QM9 dataset.}
\label{units-table}
\begin{center}
\begin{small}
\begin{sc}
\begin{tabular}{lll}
\toprule
Property   & Unit & Description    \\    
\midrule
$\mu$ & Debye    & Dipole moment\\
$\alpha$  & $Bohr^3$    & Isotropic polarizability\\
$homo$ & meV    & Energy of Highest occupied molecular orbital (HOMO)\\
$lumo$&  meV   & Energy of Lowest occupied molecular orbital (LUMO)\\
$gap$&   meV   &Gap, difference between LUMO and HOMO \\
$r2$ &  $Bohr^2$    & Electronic spatial extent\\
$ZPVE$ &   meV   & Zero point vibrational energy\\
$c_v$  &  $\frac{cal}{mol K}$    & Heat capacity at 298.15 K \\    
\bottomrule
\end{tabular}
\end{sc}
\end{small}
\end{center}
\vskip -0.1in
\end{table}

\subsection{Confidence Interval Details}\label{confidence-intervals}
\col{All the specified confidence intervals in our work are standard deviations calculated from different weight initializations using the seeds $[1,2,3,4,5,6]$ or $[1,2,3,4]$. The following tables provide additional confidence intervals for the results in the main text.}

\begin{table}[htpb]
\caption{\col{Additional confidence intervals of our method in Table \ref{qm-table}. All standard deviations are calculated from 4 seeds except for the homo property where 6 are used.}
}

\label{qm-table-confidence-intervals}
\begin{center}
\begin{small}
\begin{sc}
\setlength{\tabcolsep}{5pt}
\begin{tabular}{lc|ccc}
\toprule
&\multicolumn{1}{c}{} & \multicolumn{3}{c}{ Our 3D Infomax}\\

         Target  & \multicolumn{1}{c}{Rand Init}  & QM9 & Drugs & \multicolumn{1}{c}{QMugs}    \\    
\midrule
$\mu$       &0.4133$\pm$0.003   &\cellcolor{slight-good} \textbf{0.3507}$\pm$0.005     &\cellcolor{slight-good} \textbf{0.3512}$\pm$0.010    &\cellcolor{slight-good} 0.3668$\pm$0.004  \\
$\alpha$    &0.3972$\pm$0.014   &\cellcolor{slight-good} 0.3268$\pm$0.006              &\cellcolor{slight-good} 0.2959$\pm$0.009              &\cellcolor{slight-good} \textbf{0.2807}$\pm$0.012       \\
homo        &82.10$\pm$0.33     &\cellcolor{slight-good}\textbf{68.96}$\pm$0.32     &\cellcolor{slight-good} 70.78$\pm$0.82              &\cellcolor{slight-good} 70.77$\pm$0.74   \\
lumo        &85.72$\pm$1.62     &\cellcolor{slight-good} \textbf{69.51}$\pm$0.54     &\cellcolor{slight-good} 71.38$\pm$0.74               &\cellcolor{slight-good} 78.10$\pm$0.69    \\
gap         &123.08$\pm$3.98    &\cellcolor{slight-good} \textbf{101.71}$\pm$2.03     &\cellcolor{slight-good} 102.59$\pm$3.27             &\cellcolor{slight-good} 103.85$\pm$1.92  \\
r2          &22.14$\pm$0.21     &\cellcolor{slight-good} \textbf{17.39}$\pm$0.94       &\cellcolor{slight-good} 18.96$\pm$0.69              &\cellcolor{slight-good} 18.00$\pm$0.40   \\
ZPVE        &15.08$\pm$2.83     &\cellcolor{slight-good} 7.966$\pm$1.87               &\cellcolor{slight-good} 9.677$\pm$1.29              & \cellcolor{bit-good}12.06$\pm$2.40 \\
$c_v$       &0.1670$\pm$0.004   &\cellcolor{slight-good} 0.1306$\pm$0.009              &\cellcolor{bit-good} 0.1409$\pm$0.016                &\cellcolor{slight-good} \textbf{0.1208}$\pm$0.008  \\
\bottomrule
\end{tabular}
\setlength{\tabcolsep}{6pt}
\end{sc}
\end{small}
\end{center}
\vskip -0.1in
\end{table}

\begin{table}[htpb]
\caption{\col{Additional confidence intervals for Table \ref{drugs-table}.}}
\label{drugs-table-confidence-intervals}
\begin{center}
\begin{small}
\begin{sc}
\begin{tabular}{lll}
\toprule
Method & Gibbs & $\langle E \rangle$  \\    
\midrule
Rand Init & .2035$\pm$ 0.0011 &  .1026$\pm$ 0.0017  \\
GraphCL & .1941 & .0995   \\
3D Infomax QM9 & .1852  & .0968   \\
3D Infomax Drugs & \textbf{.1811} & \textbf{.0952} \\
3D Infomax QMugs & .1835 & .0965  \\
\bottomrule

\end{tabular}
\end{sc}
\end{small}
\end{center}
\vskip -0.1in
\end{table}

\begin{table}[htpb]
\caption{Additional confidence intervals for 3D pre-training baselines in Table \ref{quantum-properties}}
\label{predictive-comparison-table-confidence-intervals}
\begin{center}
\begin{small}
\begin{sc}
\begin{tabular}{lllllllll}
\toprule
Method    & ${\mu}$ & $\alpha$  & homo & lumo  & gap & r2 & ZPVE & $c_v$  \\ 
\midrule
Rand Init & $\pm$0.003 & $\pm$0.014 & $\pm$0.33 & $\pm$1.62 & $\pm$3.98 & $\pm$0.21 & $\pm$2.83 & $\pm$0.004    \\
3D Infomax  & $\pm$0.010 & $\pm$0.009 & $\pm$0.82 & $\pm$0.74 & $\pm$3.27 & $\pm$0.69 & $\pm$1.29 & $\pm$0.016 \\

\end{tabular}
\end{sc}
\end{small}
\end{center}
\vskip -0.1in
\end{table}

\section{Additional Results}

\subsection{Non-Quantum Properties}\label{non-quantum-experiment}
\begin{table*}[htpb]

\caption{\col{Comparison of 3D pre-training baselines and \textbf{GraphCL} against \textbf{3D Infomax}} on various OGB datasets. Shown is either the root mean squared error (RMSE) (lower is better) or the area under the ROC-curve (ROC-AUC) (higher is better). Colors indicate \textcolor{textblue}{improvement}, \textcolor{textorange}{worse} performance, or \textcolor{neutral-text}{no significant change} compared to the randomly initialized (\textbf{Rand Init}) model.
}

\label{ogb-table-main-results}

\begin{small}
\begin{center}
\begin{sc}
\begin{tabular}{llccccc}
\toprule
         Dataset   & Metric  & Rand Init & \col{DisPred} & \col{ConfGen} & GraphCL & 3D Infomax \\
\midrule
esol & RMSE $\downarrow$  & 0.947\scriptsize$\pm$0.038    & \cellcolor{bit-bad}0.986\scriptsize$\pm$0.025    & \cellcolor{slight-good}0.867\scriptsize$\pm$0.045      & \cellcolor{neutral}0.959\scriptsize$\pm$0.047        & \cellcolor{slight-good}0.894\scriptsize$\pm$0.028   \\
lipo & RMSE $\downarrow$ & 0.739\scriptsize$\pm$0.009     & \cellcolor{bit-good}0.718\scriptsize$\pm$0.021   & \cellcolor{bit-bad}0.757\scriptsize$\pm$0.035      & \cellcolor{bit-good}0.714\scriptsize$\pm$0.011       & \cellcolor{slight-good}0.695\scriptsize$\pm$0.012   \\
freesolv & RMSE $\downarrow$ & 2.233\scriptsize$\pm$0.261 & \cellcolor{bit-bad}2.486\scriptsize$\pm$0.222 & \cellcolor{bit-bad}2.428\scriptsize$\pm$0.155   & \cellcolor{slight-bad}3.744\scriptsize$\pm$0.292      &  \cellcolor{neutral}2.337\scriptsize$\pm$0.227      \\
bace & ROC-AUC $\uparrow$ &78.13\scriptsize$\pm$1.30      & \cellcolor{bit-bad}76.51\scriptsize$\pm$1.95     & \cellcolor{bit-good}80.02\scriptsize$\pm$1.58       & \cellcolor{neutral}77.18\scriptsize$\pm$4.01      & \cellcolor{bit-good}79.42\scriptsize$\pm$1.94        \\
bbbp & ROC-AUC $\uparrow$ &68.27\scriptsize$\pm$1.98      & \cellcolor{bit-bad}66.06\scriptsize$\pm$ 1.84    & \cellcolor{bit-bad}66.16\scriptsize$\pm$2.24      & \cellcolor{bit-good}71.06\scriptsize$\pm$2.00     & \cellcolor{neutral}69.10\scriptsize$\pm$1.07            \\
tox21 & ROC-AUC $\uparrow$ &73.88\scriptsize$\pm$0.64     & \cellcolor{neutral}73.87\scriptsize$\pm$0.43 & \cellcolor{bit-good}75.24\scriptsize$\pm$1.00   & \cellcolor{slight-good}78.92\scriptsize$\pm$0.61  & \cellcolor{neutral}74.46\scriptsize$\pm$0.74            \\
toxcast & ROC-AUC $\uparrow$ & 63.62\scriptsize$\pm$0.48  & \cellcolor{bit-bad}61.58\scriptsize$\pm$0.58    & \cellcolor{bit-good}64.74\scriptsize$\pm$1.20      & \cellcolor{bit-good}64.95\scriptsize$\pm$0.31     & \cellcolor{bit-good}64.41\scriptsize$\pm$0.88           \\
clintox & ROC-AUC $\uparrow$ &58.98\scriptsize$\pm$5.40   & \cellcolor{neutral}55.77\scriptsize$\pm$5.86    & \cellcolor{slight-good}64.27\scriptsize$\pm$5.22       & \cellcolor{bit-bad}51.07\scriptsize$\pm$5.52      & \cellcolor{neutral}59.43\scriptsize$\pm$3.21        \\
sider & ROC-AUC $\uparrow$ &55.95\scriptsize$\pm$3.27     & \cellcolor{neutral}57.13\scriptsize$\pm$1.89   & \cellcolor{neutral}56.34\scriptsize$\pm$4.20     & \cellcolor{neutral}57.32\scriptsize$\pm$5.00     & \cellcolor{neutral}53.37\scriptsize$\pm$3.34            \\
hiv & ROC-AUC $\uparrow$ &77.06\scriptsize$\pm$3.16       & \cellcolor{bit-bad}75.66\scriptsize$\pm$1.26     & \cellcolor{neutral}76.57\scriptsize$\pm$1.39       & \cellcolor{neutral}76.06\scriptsize$\pm$1.06      & \cellcolor{neutral}76.08\scriptsize$\pm$1.33        \\
\bottomrule
\end{tabular}
\end{sc}
\end{center}
\end{small}
\vskip -0.1in
\end{table*}

We found that 3D Infomax yields large improvements for predicting quantum properties. For non-quantum properties, there is less empirical evidence that explicit 3D information improves prediction accuracy with first explorations by \citet{conformerEnsembles} showing only little improvements for their COVID 19 related predictions. Nevertheless, for tasks such as binding prediction in the \textit{bace} dataset, we would expect it to be helpful, and we compare different methods pre-trained with GEOM-Drugs.

In Table \ref{ogb-table-main-results}, we find that 3D Infomax improved performance for 4 out of 10 OGB datasets. In contrast to  the results for quantum mechanical property predictions (Section \ref{quantum-properties}), it is not always superior to GraphCL \col{and ConfGen}. However, 3D Infomax never decreases performance which can be valuable in practice and make the method worth employing for non-quantum properties as well.

When investigating for which tasks 3D Infomax is useful, we see that abstract tasks such as predicting clinical test outcomes \col{(\textit{clintox})} benefit less.
The most significant improvements are rather possible for tasks like predicting solubility and lipophilicity in \textit{esol} and \textit{lipo}. These \col{are more directly related to molecular mechanics and a molecule's intrinsic properties (e.g., the dipole moment/polarity is important for predicting lipophilicity). They do not depend} on how a molecule will interact with others to result in, e.g., different effects on patients. \col{For such tasks, \textit{ConfGen} often leads to significant improvements, providing further evidence for the value of 3D pre-training.}

Additionally, for datasets like \textit{bace} with its binding prediction task where 3D information should be valuable, the improvements are only modest. This could suggest that our method does not capture all of the 3D information that is relevant for predicting protein binding, and there is still room for improvement. Another explanation is that a molecule's geometry is less helpful for \textit{bace} since \col{the geometry} of the protein and the binding pocket the molecule has to fit into are not known.




\subsection{Different 3D Networks and Ablation}\label{3d-network-appendix}

In this section, we justify the design of our 3D network, which we call \textit{Net3D} in this comparison. In Table \ref{3D-networks-table} we compare \textit{Net3D} with different alternative 3D networks, which are the 3D GNNs SMP and EGNN operating on learned node embeddings similar to \textit{Net3D}.
Additionally, we ablate the use of our $\gamma$ function that maps the pairwise distances to a higher dimensional space since it would be an unnecessary complication if it provides no benefit. We call it \textit{Net3D w/o $\gamma$} if \textit{Net3D} directly operates on the pairwise distances.

\begin{table}[htbp]
\caption[Comparison of 3D networks.]{Comparison of 3D networks. The MAE of the homo property pre-training and fine-tuning on different halves of QM9. \textbf{\textit{Net3D w/o $\gamma$}} refers to dropping the distance encoding of Net3D. \textit{Net3D} achieves the best MAE.}
\label{3D-networks-table}
\begin{center}
\begin{small}
\begin{sc}
\begin{tabular}{ll}
\toprule
  Method                                      & \texttt{QM9} MAE  \\    
\midrule
Rand Init                         & 82.10\scriptsize$\pm$0.33          \\
SMP                         & 72.37         \\
EGNN                         & 70.46       \\
\textit{Net3D} w/o $\gamma$                  & 70.34         \\
\textit{Net3D}                                & \textbf{68.96\scriptsize$\pm$0.32}      \\
\bottomrule
\end{tabular}
\end{sc}
\end{small}
\end{center}
\vskip -0.1in
\end{table}

In Table \ref{3D-networks-table} we can observe that \textit{Net3D} yields the best downstream performance and that using our $\gamma$ function is a valuable component of it. Possibly EGNN would benefit similarly from this encoding. SMP's downstream performance is the worst which could be expected since the 3D input representation which it uses does not uniquely define all the relative positions in a molecule.

We note that SMP is able to distinguish chiral molecules, unlike the other 3D networks, but this advantage cannot be evaluated with our experiments on quantum mechanical properties. Chirality only becomes relevant when considering the interactions between molecules, and in these situations, SMP might be able to leverage its advantage such that our evaluation could be criticized as not entirely fair. Additionally, SMP has much lower memory requirements since it does not suffer from the quadratic complexity of EGNN and \textit{Net3D} in the molecule size. Nevertheless, \textit{Net3D} performs the best, and for drug-like molecules, the quadratic complexity is not problematic.

\subsection{Different Methods for Multiple Conformers}\label{multiple-conformers-appendix}
We test three main approaches and variations of them for incorporating the 3D information of multiple conformers to justify our choice in the main text. The most straightforward one is \textbf{conformer sampling}. We use one of the single conformer setups, but when sampling the batch, we additionally sample $j\in\{1\dots c_i \}$ and use the single conformer $R_i^j$. The probability of sampling a conformer is either distributed uniformly (so $1/c_i$ is the probability for each $j$) or given by the Boltzmann weight of each conformer. 

\textbf{multi3D} is the approach from the main text where we include multiple conformers as additional positive pairs in contrastive learning. For each molecule $(G_i,\{R^j_i\}_{j\in\{1\dots c_i \}})$ we choose the $c$ lowest energy conformers to have a fixed number of them. If there are fewer than $c$ conformers for a molecule ($c_i<c$), then the lowest energy conformer is repeated. For every molecule the 3D network now takes all $c$ conformers  $\{R^j_i\}_{j\in\{1\dots c \}}$ as input and produces their latent 3D representations $\{z^b_{i,j}\}_{j\in\{1\dots c \}}$ which we can see as additional positive samples. In our contrastive setting, we, therefore, want the similarity between $z^a_i$ and all conformer representations that come from the same molecule $z^b_{i,j}$ to be high. As such, we modify \col{our loss} to obtain:
\begin{equation}
    \col{\mathcal{L}^{multi3D}} = -\frac{1}{N}\sum_{i=1}^{N}\left[log\, \frac{\sum_{j=1}^c e^{sim_{cos}(z^a_i,\,z^b_{i,j})/\tau}}{\sum_{\substack{k=1 \\ k\neq i}}^{N}\sum_{j=1}^c e^{sim_{cos}(z^a_i,\,z^b_{k,j})/\tau}} \right].
\end{equation}

One concern with this formulation is the following. Let us consider a single molecule. The objective of high similarity between the many 3D representations and the single 2D representation might be easier to solve through encoding the same 2D information in the 3D representations instead of capturing the 3D information of all conformers in the single 2D representation. The 2D network would therefore not learn to produce 3D information from its 2D inputs because the mutual information could be maximized through encoded 2D information. 

To address this problem, \textbf{multi3D+2D} is our third approach. The 2D network is now modified to produce $c$ many latent 2D representations $f^a(G_i)=\{z^a_{i,j}\}_{j\in\{1\dots c \}}$ which are compared to all 3D representations of the same molecule in a similarity function $sim$. We simply use this similarity in \col{the loss} instead of the cosine similarity. Intuitively, the 2D network now has to produce an embedding for each 3D conformer.

One way to define such a similarity between two same-sized sets of vectors is to use the sum of all pairwise cosine similarities (for brevity we drop the subscript and only write $\{z^a_{i,j}\}$ to mean the set of all representations corresponding to the i-th molecule):
\begin{equation}
    sim_{all}(\{z^a_{i,j}\},\{z^b_{i,j}\}) =  \sum_{j=1}^c \sum_{k=1}^c sim_{cos}(z^a_{i,j},\, z^b_{i,k})
\end{equation}

More principled would be to find the optimal transport matching with the highest cosine similarity, such that one 2D representation always corresponds to one 3D representation. However, this approach was not computationally feasible with the batch sizes we use in contrastive learning. We instead opt for an upper bound on the maximum similarity matching. For every 3D representation, we choose the 2D representation that has the highest similarity. This way, one 2D representation could be associated with multiple 3D embeddings, and we no longer have a mass preserving matching:
\begin{equation}
    sim_{max}(\{z^a_{i,j}\},\{z^b_{i,j}\}) =   \sum_{k=1}^c \max_{j\in\{1\dots c \}} sim_{cos}(z^a_{i,j},\, z^b_{i,k}).
\end{equation}

Beyond these similarity measures, we explore additional ones based on the inverse of different distance functions and asymmetric metrics such as the maximum mean discrepancy \cite{JMLR:v13:gretton12a} or the KL- and JS-Divergence when interpreting the conformer representations as samples from a normal distribution.

\paragraph{Results}
We evaluate which of the different approaches best leverage the additional conformer's information to justify our choice for \textit{multi3D} in the main text. Another hypothesis we wish to test is that for smaller molecules such as those in QM9, the ability to make predictions informed by multiple conformers is not as important as for larger drug-like molecules. The reasoning is that a single conformer takes most of the Boltzmann weight for QM9's molecules due to the fewer degrees of freedom. 

We test \textit{conformer sampling}, \textit{multi3D}, and \textit{multi3D+2D} when pre-training on either QMugs or one half of GEOM-Drugs and fine-tuning on QM9 or the other half of GEOM-Drugs. In QMugs we have three diverse conformers available for each molecule which are all used, while for GEOM-Drugs different numbers of conformers are available of which we use the five with the highest Boltzmann weight i.e. lowest energy. If there are fewer than five we duplicate the lowest energy conformer (see Section \ref{multiple-conformers} for details). We recall that for the \textit{multi3D+2D} loss sets with as many elements as conformers are produced by the 2D and 3D networks. Both the discussed $sim_{all}$ and $sim_{max}$ are used as similarity measures between those sets. For the \textit{conformer sampling} strategies of using a uniform weighting or sampling conformers according to their Boltzmann weight, we do not evaluate the latter on QMugs since we do not have it available with exactly three conformers per molecule.

\begin{table}[htpb]
\caption{Comparison of strategies for using multiple conformers. The middle double-column shows the results for pre-training on one half of GEOM-Drugs and the right double-column corresponds to pre-training on QMugs, and the second row indicates what dataset was used for fine-tuning. The \textbf{\textit{Random Init}} row shows the performance when training from scratch without any pre-training. For QM9, the reported number is the MAE of the homo property, and for GEOM-Drugs it is the MAE when predicting the ensemble Gibbs free energy. There are large improvements from using multiple conformers, but the differences between the methods are small.}
\label{3D-methods-table}
\begin{center}
\begin{small}
\begin{sc}
\begin{tabular}{l|ll|ll}
\toprule
\multirow{2}{*}{Loss/Estimator } &\multicolumn{2}{c|}{\texttt{GEOM-Drugs} pre-training} & \multicolumn{2}{c}{\texttt{QMugs} pre-training}\\
& \texttt{QM9} & \texttt{GEOM-Drugs} & \texttt{QM9}  & \texttt{GEOM-Drugs}   \\    
\midrule
Rand Init                                & 82.10\scriptsize$\pm$0.33      & .2035\scriptsize$\pm$.0011    & 82.10\scriptsize$\pm$0.33    & .2035\scriptsize$\pm$.0011 \\
single conformer                                 & 71.66    &  .1844    &   82.57   & .1966  \\
uniform sampling                                 & \textbf{70.66}    &  \textbf{.1823}     &  72.94        & .1874  \\
boltzmann sampling                              &  \textbf{70.93}    &  .1846     &  x       & x \\
multi3D                                         & \textbf{70.78}     & \textbf{.1811}     &  \textbf{70.77}        & \textbf{.1831} \\
$sim_{all}$                                     &  71.11    &  .1849     &  72.40       & .1936 \\
$sim_{max}$                                     &  \textbf{70.81}    &  .1896     &   \textbf{71.15}       & \textbf{.1840} \\
\bottomrule
\end{tabular}
\end{sc}
\end{small}
\end{center}
\vskip -0.1in
\end{table}

In Table \ref{3D-methods-table} we can observe that there are large improvements possible when using multiple conformers. After pre-training on QMugs, the MAE, when predicting the homo property, decreases from $82.57$ to $70.77$ and from $.1966$ to $.1831$ for predicting the Gibbs free energy for GEOM-Drugs. Notably, these improvements are much larger than when pre-training with GEOM-Drugs. This is likely because the GEOM-Drugs dataset contains the lowest energy conformers, and we always use the most probable one with the highest Boltzmann weight when pre-training with a single conformer.
Meanwhile, the QMugs dataset contains three diverse conformers per molecule and not the ones with the highest Boltzmann weight. Pre-training with the lowest energy conformer from GEOM-Drugs already captures most of the relevant information, and using more is not as beneficial. However, for QMugs, using the information of all three diverse conformers is crucial.

Similar to the small improvements over the random initialization baseline with GEOM-Drugs, the different methods for using multiple conformers mostly perform the same when pre-training with GEOM-Drugs. When pre-training with QMugs instead, the MAE are overall slightly worse, and we find \textit{multi3D} to perform the best. Note that this is with the slight caveat that the epoch at which pre-training is stopped for all methods was chosen based on where \textit{multi3D} had the lowest MAE.

Due to these results, we consider \textit{multi3D}, and \textit{conformer sampling} with uniform weighting as our best methods since \textit{multi3D} performs slightly better with pre-training on QMugs but \textit{conformer sampling} is simpler and especially uses much less memory. For \textit{multi3D}, all the conformers need to be processed in parallel, and training with more than 5 conformers and a batch size of 500 would not be possible on a 48GB vRAM GPU.

The hypothesis that the downstream performance on the smaller molecules of QM9 would benefit less from using multiple conformers than the molecules of GEOM-Drugs clearly does not hold. Surprisingly, the improvements on the small molecules of QM9 are larger.

\subsection{Different Losses}\label{losses-experiments}
We compare the different losses to estimate and maximize the mutual information. For this purpose, we pre-train PNA on $50\,000$ molecules from QM9 and another instance on $140\,000$ molecules of GEOM-Drugs, both with a single conformer. We do so with the Donsker-Varadhan \cite{DIM} estimator, the Jensen-Shannon estimator \cite{DIM}, InfoNCE, and our loss. For our loss, we search over seven temperature parameters $\tau \in [0.05, 0.1, 0.2, 0.3, 0.4, 0.5, 0.7]$ and choose $\tau=0.1$.

\begin{table}[htpb]
\caption{Comparison of mutual information estimators for 3D Infomax. The middle double-column shows the results for pre-training on one half of QM9, and the right double-column corresponds to pre-training on one half of GEOM-Drugs, and the second row indicates what dataset was used for fine-tuning. The \textbf{\textit{Rand Init}} row shows the performance when training from scratch without any pre-training. For QM9 the reported number is the MAE of the homo, and for GEOM-Drugs it is the MAE when predicting the ensemble Gibbs free energy.}
\label{losses-table}
\begin{center}
\begin{small}
\begin{sc}
\begin{tabular}{l|ll|ll}
\toprule
 \multirow{2}{*}{Loss/Estimator } &\multicolumn{2}{c|}{\texttt{QM9} pre-training} &  \multicolumn{2}{c}{\texttt{GEOM-Drugs} pre-training}\\

                                        & \texttt{QM9} & \texttt{GEOM-Drugs} & \texttt{QM9}  & \texttt{GEOM-Drugs}   \\    
\midrule
Rand Init                                 & 82.10\scriptsize$\pm$0.33      & .2035\scriptsize$\pm$.0011    & 82.10\scriptsize$\pm$0.33    & .2035\scriptsize$\pm$.0011 \\
Donsker-Varadhan                            & 82.49                 & .2152    & 85.46    & .2013 \\
Jensen-Shannon                              & 80.71                 & .2078   &  81.61   & .2047 \\
InfoNCE                               & 75.81                 & \textbf{.1938}    &  79.31   & .1894 \\
our loss                                & \textbf{68.96\scriptsize$\pm$0.32}      & \textbf{.1945}    &  \textbf{71.66}   & \textbf{.1844} \\
\bottomrule
\end{tabular}
\end{sc}
\end{small}
\end{center}
\vskip -0.1in
\end{table}

In Table \ref{losses-table} we see that 3D pre-training on GEOM-Drugs or QM9 can yield significant improvements for predicting quantum mechanical properties, especially when using InfoNCE and our loss as objectives. These two objectives perform better than the Donsker-Varadhan, and Jensen-Shannon estimator in every case and the Jensen-Shannon objective is superior to the Donsker-Varadhan estimator, which seems to yield no significant improvements over random initialization. The superiority of the Jensen-Shannon loss over the Donsker-Varadhan alternative is in line with the findings of \citet{DIM} in their different setting on images.
While our loss seems to perform better than InfoNCE in three settings, this might be due to the additional investment in searching through temperature parameters for our loss. 

\subsection{SSL Methods}\label{latent-space-ssl}
Here we compare our 3D Infomax pre-training against three additional SSL methods. These are Barlow Twins  \cite{BarlowTwins}, multi-modal BYOL \cite{BYOL}, and VICReg \cite{vicreg}. We pre-train these methods on one half of QM9. For a fair comparison, we search through 8 different hyperparameter settings based on the downstream performance on the QM9 homo property. After these method-specific hyperparameters were selected, we tuned every method with a random search over the same search space.

For 3D Infomax, we vary the temperature of our loss $\tau$. When using BYOL we try different decay rates $\gamma$ for the exponential moving average weight copying. Here, we include $\gamma=0$ making our setup similar to a multi-modal version of SimSiam \cite{SimSiam}. For Barlow Twins, the hyperparameter is $\lambda$ weighting the redundancy loss. Lastly, for VICReg we vary $\mu$ and $\nu$, the parameters for the variance and the covariance regularization:
\begin{enumerate}
    \item 3D Infomax with our loss: $\tau \in [0.02, 0.05, 0.1, 0.2, 0.3, 0.4, 0.5, 0.7]$ where $\tau=0.01$ performed the best.
    \item Multi-modal BYOL: $\gamma \in [0, 0.0005, 0.001, 0.005,0.01, 0.03, 0.05, 0.07]$ where $\gamma = 0.005$ performed the best.
    \item Barlow-Twins: $\lambda \in [0.002, 0.0039, 0.005, 0.007, 0.01, 0.012, 0.015, 0.02]$ where $\gamma = 0.0039$ performed the best.
    \item VICReg: $\lambda = 1$; $\mu \in [1, 0.5]$; $\nu \in [0.02, 0.04, 0.1, 0.3]$ where $\lambda=1, \mu=1, \nu=0.04$ performed the best
\end{enumerate}

\begin{table}[htpb]
\caption{Comparison of latent space SSL methods. The numbers show the MAE when predicting QM9's homo property after pre-training on one half of QM9 with the given method and fine-tuning on the other half of QM9. The \textbf{\textit{Rand Init}} column shows the MAE without pre-training and with random weight initialization. 3D Infomax is our best latent space SSL method.}
\label{ssl-methods-experiment-table}
\begin{center}
\begin{small}
\begin{sc}
\begin{tabular}{l|lllll}
\toprule
 & Random Init & 3D Infomax & BYOL & Barlow Twins & VICReg \\
\midrule
QM9 MAE     & 82.10\scriptsize$\pm$0.33      & \textbf{68.96\scriptsize$\pm$0.32}   & 79.16\scriptsize$\pm$0.58    &  82.38\scriptsize$\pm$0.48 &  85.15 \\
\bottomrule

\end{tabular}
\end{sc}
\end{small}
\end{center}
\vskip -0.1in
\end{table}

The results in Table \ref{ssl-methods-experiment-table} showcase that 3D Infomax clearly is the superior method in our setting. It decreases the MAE from 82.10 $\pm$ 0.33 to $68.96 \pm 0.32$
while the other methods either lead to no improvement or to the much smaller drop to 79.16 $\pm$ 0.58 for BYOL. This is not due to collapse to a constant solution since we can observe a high variance between the representations in a batch for all methods. Furthermore, with the final parameter settings, all methods were able to achieve a low value for their loss during pre-training, both on the training and validation data and there are no optimization issues. 

Intuitively, the results can be explained by 3D Infomax being the only method that optimizes a lower bound on the mutual information, which potentially makes it especially fit for our setting. The other approaches have no direct relation to mutual information and instead rely on maximizing a notion of similarity with tricks to prevent collapse. While this might work for conventional SSL, we see no success in our scenario where the rigorous guarantee on maximizing the mutual information seems valuable.

Another reason for the poor performance of BYOL and especially Barlow Twins and VICReg might be that they rely on having symmetric networks to generate the compared representations. In our scenario, we have very little similarity between the architectures with our 2D and 3D networks operating on different modalities. This hypothesis would fit in line with the findings of \citet{vicreg} and \citet{BarlowTwins} where introducing asymmetries between the networks hurt performance.

\subsection{\col{Pre-training a 3D GNN}}
\col{We try to use our 3D Infomax setup to pre-train a 3D GNN. For this purpose, we employ SMP \cite{sphericalmessagepassing2021} as 3D network during pre-training with half of the QM9 dataset. We then transfer it's weights and fine tune them using the accurate 3D conformers of the other half of QM9's molecules to predict the dataset's properties. We compare this with SMP trained on the same molecules with randomly initialized weights. The only architectural difference between the networks is that the pre-trained GNN does not use atom features for the reasons explained in the 3D Network paragraph in Section \ref{method}.}

\begin{table}[htpb]
\caption{\col{
MAE for predicting QM9's molecular properties. SMP is tested with random weight initialization and with the weights obtained from using it as 3D network in our 3D Infomax pre-training setup.}
}
\label{pre-train-SMP-table}
\begin{center}
\begin{small}
\begin{sc}
\setlength{\tabcolsep}{5pt}
\begin{tabular}{lcc}
\toprule
         Target  & SMP Rand Init & SMP pre-trained   \\    
\midrule
$\mu$       &0.0726    &0.0801  \\
$\alpha$    &0.1542    &0.1276  \\
homo      &56.19      &44.50    \\
lumo      &43.58     & 37.48   \\
gap       &85.10     &70.45  \\
r2        &1.51      &1.12   \\
ZPVE      &2.69     &2.43    \\
$c_v$       &0.0498    & 0.0421 \\
\bottomrule
\end{tabular}
\setlength{\tabcolsep}{6pt}
\end{sc}
\end{small}
\end{center}
\vskip -0.1in
\end{table}

\col{In Table \ref{pre-train-SMP-table}, we find that pre-training improves the 3D GNN's performance. This may be due to the covalent bonding structure and other 2D edge information that is available during pre-training and which SMP usually cannot use since it employs a radius graph. This is the case even though the pre-trained SMP does not have access to the atom features. Pre-training 3D GNNs might be an interesting future direction to attempt beating the state-of-the-art methods for predicting quantum properties with accurate 3D information.}

\subsection{Cheap Neural Conformers as 3D GNN input}\label{geomol-conformers}

\col{In Section \ref{quantum-properties} we used RDKit's ETKDG algorithm \cite{rdkit} to generate inaccurate but cheap and fast conformers and employed them as inputs to the 3D GNN SMP \cite{sphericalmessagepassing2021}. Here, we attempt the same with conformers generated by the state-of-the-art deep learning method for conformation generation which is GeoMol \cite{ganea2021geomol}. For this purpose, we train GeoMol with $50$k molecules of QM9 and use it to generate the conformations for the rest of QM9. SMP is then trained on $50$k different molecules to predict their properties, either using RDKit's conformers or those of GeoMol. This enables a fair comparison with 3D Infomax, which uses the same molecules for pre-training that were used to train GeoMol. When visually inspecting some of the conformers generated by GeoMol, we found that they were sometimes of poor quality for molecules with rings and contained outliers with conformations that seem particularly unrealistic.}

\begin{table}[htpb]

\caption{\col{MAE for QM9's properties. \textbf{3D Infomax} is tested with three pre-training datasets and compared with the 3D GNN SMP using explicit 3D coordinates. The conformers are generated using the classical method RDKit ETKDG or the learned method GeoMol. Colors indicate \textcolor{textblue}{improvement} (lower MAE) or }\textcolor{textorange}{worse} performance compared to the randomly initialized (\textbf{Rand Init}) model.
}

\label{geomol-conformers-table}
\begin{center}
\begin{small}
\begin{sc}
\setlength{\tabcolsep}{5pt}
\begin{tabular}{lc|ccc|cc}
\toprule

&\multicolumn{1}{c}{}  &  \multicolumn{3}{c}{ Our 3D Infomax}& \multicolumn{1}{c}{\col{RDKit}} & {\col{GeoMol}}\\

         Target  & \multicolumn{1}{c}{Rand Init}  & QM9 & Drugs & \multicolumn{1}{c}{QMugs} & \multicolumn{1}{c}{\col{SMP}} & \multicolumn{1}{c}{\col{SMP}}   \\    
\midrule
$\mu$       &0.4133\scriptsize$\pm$0.003 &\cellcolor{slight-good} \textbf{0.3507}    &\cellcolor{slight-good} \textbf{0.3512}    &\cellcolor{slight-good} 0.3668 &\cellcolor{slight-bad} 0.4344   & \cellcolor{slight-bad}0.6046   \\
$\alpha$    &0.3972\scriptsize$\pm$0.014 & \cellcolor{slight-good} 0.3268    &\cellcolor{slight-good} 0.2959    &\cellcolor{slight-good} \textbf{0.2807}   &\cellcolor{slight-good} 0.3020 & \cellcolor{slight-bad}0.8435   \\
homo      &82.10\scriptsize$\pm$0.33    &\cellcolor{slight-good}\textbf{68.96}     &\cellcolor{slight-good} 70.78     &\cellcolor{slight-good} 70.77  &\cellcolor{bit-bad} 82.51    & \cellcolor{slight-bad}195.0  \\
lumo      &85.72\scriptsize$\pm$1.62     &\cellcolor{slight-good} \textbf{69.51}     &\cellcolor{slight-good} 71.38     &\cellcolor{slight-good} 78.10   &\cellcolor{bit-good} 80.36  & \cellcolor{slight-bad}201.4 \\
gap       &123.08\scriptsize$\pm$3.98     &\cellcolor{slight-good} \textbf{101.71}    &\cellcolor{slight-good} 102.59    &\cellcolor{slight-good} 103.85 &\cellcolor{bit-good} 114.24  & \cellcolor{slight-bad}284.1 \\
r2        &22.14\scriptsize$\pm$0.21     &\cellcolor{slight-good} \textbf{17.39}     &\cellcolor{slight-good} 18.96     &\cellcolor{slight-good} 18.00  &\cellcolor{bit-bad} 22.63   & \cellcolor{slight-bad}65.84  \\
ZPVE      &15.08\scriptsize$\pm$2.83      &\cellcolor{slight-good} 7.966     &\cellcolor{slight-good} 9.677     & \cellcolor{bit-good}12.06 & \cellcolor{slight-good} \textbf{5.18}   & \cellcolor{slight-bad}17.40 \\
$c_v$       &0.1670\scriptsize$\pm$0.004    &\cellcolor{slight-good} 0.1306    &\cellcolor{bit-good} 0.1409    &\cellcolor{slight-good} \textbf{0.1208} & \cellcolor{bit-good} 0.1419   & \cellcolor{slight-bad}0.5467  \\
\bottomrule
\end{tabular}

\setlength{\tabcolsep}{6pt}
\end{sc}
\end{small}
\end{center}
\vskip -0.1in
\end{table}

\col{Table \ref{geomol-conformers-table} shows that SMP performs poorly with the conformers generated by GeoMol and using those generated by RDKit is always superior. This is the case even though the average accuracy of GeoMol's conformers is comparable to that of RDKit ETKDG's conformers when GeoMol is trained on all of QM9 \cite{ganea2021geomol}. We hypothesize that the high MAE with GeoMol's conformers occur since they contained some particularly unrealistic outlier conformations, and SMP is not able to handle those well.}

\subsection{Batch Size}
It is well known that contrastive learning with losses such as NTXent heavily rely on a high number of negative samples - that is large batch sizes. Thus, we evaluate how our loss benefits from large batch sizes in Table \ref{ssl-batchsize-table}. We find that our large batch size of 500 is indeed necessary for the good performance of 3D Infomax.

\begin{table}[htpb]
\caption[Downstream performance for different pre-training batch sizes.]{Downstream performance for different pre-training batch sizes. Shown is the MAE when fine-tuning on the homo property of one half of QM9 after pre-training on the other half of QM9. }
\label{ssl-batchsize-table}
\begin{center}
\begin{small}
\begin{sc}
\begin{tabular}{lccccc}
\toprule
Batch Size & 500 & 400 & 300 & 200 & 100 \\
   
\midrule
QM9 MAE     & 68.96\scriptsize$\pm$0.32      & 69.81   & 70.18    &  71.13 &  73.65\\
\bottomrule
\end{tabular}
\end{sc}
\end{small}
\end{center}
\end{table}

\subsection{Combining Pre-Training Methods}\label{combine-pre-training}

Here we simply use GraphCL's node drop augmentation for the 2D graph and the 3D information (removing all pairwise distances for a removed atom) with a drop ratio of 0.2 during our 3D pre-training process.
\begin{table}[htpb]

\caption{Comparison of performance when combining 3D pre-training with conventional pre-training by randomly dropping nodes on the 2D or 3D side (labeled \textbf{3D Infomax + }) for various biophysical property OGB datasets. \textbf{GraphCL} is another pre-trained baseline. Shown is either the RMSE indicated by $\downarrow$ where lower values are better or the ROC-AUC indicated by $\uparrow$ where higher values are better.  Colors indicate \textcolor{textblue}{improvement}, \textcolor{textorange}{worse} performance, or \textcolor{neutral-text}{no significant change} compared to the randomly initialized (\textbf{Rand Init}) model. 3D Infomax is either on par with random initialization or better. There is no negative transfer as there is with GraphCL.}

\label{ogb-table-combining-pretraining}
\begin{center}
\begin{small}
\begin{sc}
\begin{tabular}{lcccc}
\toprule

         dataset     & Rand Init  & GraphCL & 3D Infomax & 3D Infomax +\\
\midrule
esol$\downarrow$ & 0.947\scriptsize$\pm$0.038         & \cellcolor{neutral}0.959\scriptsize$\pm$0.047        & \cellcolor{slight-good}0.894\scriptsize$\pm$0.028 & \cellcolor{bit-good}0.918\scriptsize$\pm$0.037 \\
lipo$\downarrow$& 0.739\scriptsize$\pm$0.009          & \cellcolor{bit-good}0.714\scriptsize$\pm$0.011       & \cellcolor{slight-good}0.695\scriptsize$\pm$0.012 &  \cellcolor{bit-good}0.710\scriptsize$\pm$0.007 \\
freesolv$\downarrow$& 2.233\scriptsize$\pm$0.261     & \cellcolor{slight-bad}3.744\scriptsize$\pm$0.292      &  \cellcolor{neutral}2.337\scriptsize$\pm$0.227    & \cellcolor{bit-bad}2.791\scriptsize$\pm$0.323 \\
bace$\uparrow$ &78.13\scriptsize$\pm$1.30                & \cellcolor{neutral}77.18\scriptsize$\pm$4.01      & \cellcolor{bit-good}79.42\scriptsize$\pm$1.94  &     \cellcolor{bit-good}79.28\scriptsize$\pm$3.61  \\
bbbp$\uparrow$ &68.27\scriptsize$\pm$1.98                & \cellcolor{bit-good}71.06\scriptsize$\pm$2.00     & \cellcolor{neutral}69.10\scriptsize$\pm$1.07      &  \cellcolor{neutral}68.64\scriptsize$\pm$2.19    \\
tox21$\uparrow$ &73.88\scriptsize$\pm$0.64               & \cellcolor{slight-good}78.92\scriptsize$\pm$0.61  & \cellcolor{neutral}74.46\scriptsize$\pm$0.74      &  \cellcolor{neutral}73.73\scriptsize$\pm$0.69   \\
toxcast$\uparrow$ & 63.62\scriptsize$\pm$0.48            & \cellcolor{bit-good}64.95\scriptsize$\pm$0.31     & \cellcolor{bit-good}64.41\scriptsize$\pm$0.88     &  \cellcolor{neutral}63.95\scriptsize$\pm$0.38 \\
clintox$\uparrow$ &58.98\scriptsize$\pm$5.40             & \cellcolor{bit-bad}51.07\scriptsize$\pm$5.52      & \cellcolor{neutral}59.43\scriptsize$\pm$3.21      &  \cellcolor{slight-good}83.59\scriptsize$\pm$3.64 \\
sider$\uparrow$ &55.95\scriptsize$\pm$3.27               & \cellcolor{bit-good}57.32\scriptsize$\pm$5.00     & \cellcolor{neutral}53.37\scriptsize$\pm$3.34      & \cellcolor{slight-good}58.43\scriptsize$\pm$1.28   \\
hiv$\uparrow$ &77.06\scriptsize$\pm$3.16                 & \cellcolor{bit-bad}76.06\scriptsize$\pm$1.06      & \cellcolor{neutral}76.08\scriptsize$\pm$1.33      &  \cellcolor{bit-bad}75.38\scriptsize$\pm$0.95\\
\bottomrule
\end{tabular}
\end{sc}
\end{small}
\end{center}
\vskip -0.1in
\end{table}

\end{document}